%% file: main.tex
\documentclass{sigkddExp}
\usepackage{verbatim}
\usepackage{amssymb}
\usepackage{pifont}
\newcommand{\cmark}{\ding{51}}%
\usepackage{xcolor}
\usepackage{multirow}
\usepackage{array}
\usepackage{calc}
\usepackage{relsize}
\usepackage{microtype}
\usepackage{amsthm}
\usepackage{graphicx}
\usepackage{float}
\usepackage{booktabs}

\usepackage[backend=bibtex]{biblatex}
\DeclareMathOperator*{\argmin}{arg\,min}

\newtheorem{definition}{Definition}
\newtheorem{assumption}{Assumption}
\makeatletter
\newcommand*{\centernot}{%
  \mathpalette\@centernot
}
\def\@centernot#1#2{%
  \mathrel{%
    \rlap{%
      \settowidth\dimen@{$\m@th#1{#2}$}%
      \kern.5\dimen@
      \settowidth\dimen@{$\m@th#1=$}%
      \kern-.5\dimen@
      $\m@th#1\not$%
    }%
    {#2}%
  }%
}
\makeatother

\newcommand{\indep}{\perp \!\!\! \perp}

\newcommand{\notindependent}{\centernot{\indep}}
\usepackage{amsmath}
\usepackage{multirow}

\begin{document}
%

\title{Causal Inference for Time series Analysis: Problems, Methods and Evaluation}
%

%


\author{Raha Moraffah$^*$, Paras Sheth$^*$, Mansooreh Karami$^*$, Anchit Bhattacharya$^*$ \\ Qianru Wang$^{\ddagger\ddagger}$, Anique Tahir$^*$, Adrienne Raglin$^{\dagger}$, Huan Liu$^*$\\\affaddr{$^*$Computer Science \& Engineering, Arizona State University, Tempe, AZ, USA}\\\affaddr{$^{\dagger\dagger}$Northwestern Polytechnical University}, \affaddr{$^{\dagger}$Army Research Lab, USA} \\\affaddr{$^*$\{rmoraffa, psheth5,mkarami, abhatt22, artahir, huanliu\}@asu.edu
}, $^{\dagger\dagger}$qr369wang@gmail.com,\\$^{\dagger}$adrienne.raglin2.civ@mail.mil}

\maketitle
\begin{abstract}

Time series data is a collection of chronological observations which are generated by several domains such as medical and financial fields. Over the years, different tasks such as classification, forecasting and clustering have been proposed to analyze this type of data. Time series data has been also used to study the effect of interventions overtime. Moreover, in many fields of science, learning the causal structure of dynamic systems and time series data is considered an interesting task which plays an important role in scientific discoveries. Estimating the effect of an intervention and identifying the causal relations from the data can be performed via causal inference. Existing surveys on time series discuss traditional tasks such as classification and forecasting or explain the details of the approaches proposed to solve a specific task. In this paper, we focus on two causal inference tasks, i.e., treatment effect estimation and causal discovery for time series data and provide a comprehensive review of the approaches in each task. Furthermore, we curate a list of commonly used evaluation metrics and datasets for each task and provide an in-depth insight. These metrics and datasets can serve as benchmark for research in the field.
\end{abstract}

\section{Introduction}
\input{intro.tex}

\input{TimeSeriesModelling/timeSeriesModelingNew}

\input{CausalBackground/Background}

\input{CTEEstimation/CTE-Intro}

\subsection{Causal Discovery for Time Series}
\label{discovery}


One of the fundamental tasks in any field of science is to identify the causal relations between different phenomena in a system \cite{guo2018survey, peters2017elements,spirtes2016causal}. Researchers are often interested in discovering what causes a phenomenon and how manipulating a phenomenon affects the others. 
Recently, there has been a proclivity towards creating algorithms for causal discovery on time series data \cite{xu2019scalable, hyttinen2016causal, lowe2020amortized}. Causal discovery on time series data is an important task and is often used in different fields of research. For example, determining the causal relations between the aggregated daily stock price and trading volumes or discovering how patient’s records and the prescription of a specific drug over time are related to each other are types of questions which can be answered by performing causal discovery on time series data. 
In this section, we provide a comprehensive survey of the frameworks proposed for this task by categorizing them into three main types namely Granger causality and conditional independence based approaches, structural equation based models and deep learning based frameworks and introducing the methods in each category.

\subsubsection{Methods based on Granger Causality and Conditional Independence}
Granger causality \cite{granger1969investigating} is a popular concept of causality which has been widely used to infer the causal relationships from time series data \cite{arnold2007temporal, ding200617, hiemstra1994testing}. The idea behind Granger causality is that $Y$ Granger causes $X$ if it contains some unique information about $X$ which is not available in $X$'s past as well as all the information in the universe.  
In practice, this idea is materialized by investigating whether the prediction of the current value of time series $X$ improves by incorporating $Y$'s past into its own past. If so, it is  reported that $Y$ Granger causes $X$ and has a causal influence on it. Vector autoregressive models (VAR), are common ways of modeling this problem:

\begin{equation*}
\label{granger}
X_t = \sum_{\tau=1}^{\tau_{max}} \phi(\tau) X_{t-\tau} + e_t,
\end{equation*}
where  $X_t = (X_{1t}, ... , X_{nt} )$ indicates time series $X$ at time step t, $\phi(\tau)$ is the $N \times N$ coefficient
matrix at lag $\tau$ , $\tau_{max}$  denotes the maximum time lag, and $e$ represents an independent noise. Using this equation, we say $X_i$ Granger causes $X_j$ with lag $\tau$ if any of the coefficients in $\phi_{ji}(\tau)$ is non-zero. This relationship can be shown by $X_{t-\tau}^i \rightarrow X_{t}^j$  which demonstrates the causal link between $ X_i$ and $ X_j$ at lag $\tau$.

Several VAR based frameworks have been proposed over the years to perform causal discovery on time series. For instance, In \cite{gregorova2015learning}, authors focus on linear vector autoregressive models along with stationary time series. They claim that adopting VAR model directly to identify the causal relations neglects the possibility of shared structures in the lagged dependencies captured by the causal graph. To address this issue, they propose two new methods based on multi-task learning paradigms and techniques of structured regularizations for learning the G-causality in VARs with leading indicators. The difference between the two proposed methods is that they follow different structural assumptions for the G-causality graphs. The first method called SingleCluster VAR (SCVAR)  assumes that the leading indicators within the system help predict all the series in the system . This method therefore aims at identifying such indicators; The second method named MultiCluster VAR (MCVAR) assumes that there are different indicators for different cluster of series and aims to learn the leading indicators as well as the unknown clusters.

Shojaie and Michailidis \cite{shojaie2010discovering} propose a method for estimating the causal relationships of the time series for the case where number of components $(p)$ is very large  compared to the sample size $(n)$. The authors claim that in such cases penalized methods provide higher accuracy. Based on this hypothesis, an extension of the lasso penalty named as truncating lasso penalty is proposed. This framework has two key features. First, it automatically determines the order of the Vector Autoregressive (VAR) models and secondly it performs model simplification by reducing the number of covariates in the model.
The truncating lasso estimate of the graphical Granger model can be found by solving the following estimation problem for $i = 1, ..., p$:

\begin{equation}\begin{array}{c}
\underset{\theta^{t} \in \mathbb{R}^{p}}{\operatorname{argmin}} n^{-1}\left\|\mathcal{X}_{i}^{T}-\sum_{t=1}^{d} \mathcal{X}^{T-t} \theta^{t}\right\|_{2}^{2}+\lambda \sum_{t=1}^{d} \Psi^{t} \sum_{j=1}^{p}\left|\theta_{j}^{t}\right| w_{j}^{t} \\
\Psi^{1}=1, \quad \Psi^{t}=M^{I\left\{\left\|A^{(t-1)}\right\|_{0}<p^{2} \beta /(T-t)\right\}}, t \geq 2
\end{array}\end{equation}

Where $\mathcal{X}_{i}^{t}$ is the $i$-th column of the design matrix corresponding to $t$-th time point, M is a large constant, $\beta$ is the allowed false negative rate set by the user and $\theta$ is the model parameters. 

In \cite{siggiridou2015granger}, authors propose a framework to estimate the coefficients of VAR model when there are many observed variables and short time series. To tackle this scenario, the authors propose a new dimension reduction approach designed for time series. They modify the backward-in-time-selection (BTS) approach for Granger causality. While the original BTS method includes all lags up to the selected order $p_{k}$ for each time series, $X_{k}$, the proposed modification includes only the lags of each $X_{k}$ that are selected at each step of the algorithm.

In \cite{amornbunchornvej2019variable} authors claim that most of the existing works in inferring causal relations from time series data using Granger causality assume that the lag between a cause and an effect is at a fixed time point. To address this problem, they propose a novel method that uses Dynamic Time Warping (DTW) which is a distance measure between two time series along with Granger causality to identify the variable lagged based causality in time series.

Haufe et al. \cite{haufe2010sparse} introduce a framework which aims to estimate causal interactions for multivariate time series. The authors argue that, it is more practical to find all potential causal relations between all time series at once, rather than finding the causal relations for each time series pair. They propose a framework that accounts for the fact that there is no causal relation between $z_i$ and $z_j$ if all the AR coefficients for this certain pair of time series are jointly zero. They propose a sparsification technique through statistical testing.

Another way to recover the causal dependencies from times series data is by testing conditional independence relations between variables and their pasts. Conditional independence based frameworks allow the causal graph identification under the assumption of time-order, Causal Sufficiency, Causal Markov Condition and Faithfulness.
For example, transfer entropy \cite{schreiber2000measuring} is an information theoretic approach for causal discovery which can also be considered as the generalized  Granger causality \cite{barnett2009granger}. This method checks for conditional independence between $X_{t-\tau}^i$ and $X_t^j$ given the past $(X_{t-1}, ... ,X_{t-\tau} )$. One major drawback of this method is that it highly suffers from curse of dimensionality. To alleviate this issue, Sun and Bollt \cite{sun2014causation} and Sun et al. \cite{sun2015causal} propose a framework based on optimal causation entropy principle. The framework identifies the lagged parents using an iterative procedure which consists of a forward and a backward phase.

Another conditional independence based algorithm for causal discovery is PC algorithm \cite{spirtes1991algorithm}. Even though the original PC algorithm was proposed for general random variables and did not consider the time order of the them, the algorithm can be extended to the scenario where variables are collected chronologically. Entner and Hoyer propose to adopt the Fast Causal Inference (FCI) \cite{Spirtes:2000aa}, originally designed for non-temporal data, to infer the causal relations from time series data  in the presence of unobserved variables \cite{entner2010causal}. The advantage of the proposed method over Granger causality is that it also takes the latent variables in to account while identifying causal relations. 
Recently, a variation of PC algorithm, namely PCMCI algorithm \cite{runge1702detecting} has also been proposed which consists of two steps. In the first step, the algorithm adopts PC algorithm to identify parents of $X_t^j$ ($P(X_t^j)$) for all $X_t^j \in X_t$. It then applies a momentary conditional independence (MCI) test.
In \cite{runge2020discovering} the authors propose an extension to the PCMCI algorithm called PCMCI$^+$ that includes discovery of contemporaneous links along with lagged links.

Chu et al. \cite{chu2008search} propose a causal discovery algorithm based on conditional independence which is specifically designed for nonlinear and stationary time series. The authors introduce two classes of conditional independence tests based on additive model regression to recover the structure of additive non linear time series consistently. 

Jangyodsuk et al. \cite{jangyodsuk2014causal}, propose a novel approach for causal discovery in time series data based on Pearl's causality where the causal relationships are based on Directed Acyclic Graphs (DAGs) and the conditional dependencies between the variables. The authors state that the mutual information between an effect and a cause can be incrementally constructed based on the mutual information values between the effect $Y$ and the previously found cases $X_{1:i-1}$. The final output of this method is a causal graph with each time series as a node and the edge weight for each edge is the leading time. Leading time is the difference in time steps between the occurrence of cause and the occurrence of the effect. The equation for mutual information chain rule is as shown: 

\begin{equation*}I\left(X_{i: N} ; Y\right)=\sum_{i=i}^{N} I\left(X_{i} ; Y \mid X_{1: i-1}\right)\end{equation*}

where $Y$ is the effect and $X_i: N$ are its N causes.

 \subsubsection{Methods Based on Structural Equation Models}
Structural Equation Models (SEM) have been long used to perform causal discovery from observational data \cite{zhang2006extensions, shimizu2006linear, cai2018causal, hoyer2008nonlinear}. One popular form of SEM used for identifying the causal structure is Linear Non-Gaussian Acyclic Model  (LiNGAM) \cite{shimizu2006linear}. This model makes use of Independent Component Analysis (ICA) to identify causal relations in observational data.
Unlike Gaussian Processes, LiNGAM is based on using the non gaussianity of the data. The key aspect of this method is that
it is possible to identify more of the generating structure when the data is non-gaussian.
It can be mathematically represented as :
\begin{equation}\label{LiNGAM}
    x_i = \sum_{k(j)<k(i)}b_{ij}x_j + e_i
\end{equation}
where $e_i$ are continuous latent variables that are exogenous and $b_{ij}$ are the connection strengths between variables $x_i$ and $x_j$ and k(i) denotes the causal ordering of $x_i$. The exogenous variables, $e_i$ follow Non-Gaussian distributions. In this section, We briefly introduce structural equation based methods for causal discovery specifically designed for time series data.



In \cite{hyvarinen2008causal}, Hyv{\"a}rinen et al. propose a generalized version of LiNGAM which can be considered as a combination of autoregressive models and non-Gaussian instantaneous models as defined in equation below: 

\begin{equation}\label{tslingam}
    x(t) = \sum_{\tau=0}^k B_\tau x(t-\tau) _ e(t)
\end{equation}

where $e_i(t)$ are random processes which model disturbance. This model allows estimating instantaneous effects (i.e., $X_{t}^i \rightarrow X_{t}^j$) as well as lagged effects which is shown to lead to a better estimation of causal structure. The parameters of the model are estimated using a 3-step process by estimating an autoregressive model, computing the residuals and performing LiNGAM analysis on them. The paper also shows that the model is identifiable.

Schaechtle et al. \cite{schaechtle2013multi} propose to integrate Linear Non-Gaussian Acyclic Model (LiNGAM) with tensor analytic techniques to identify causal relationships from the high dimensional data. 

Rothenh{\"a}usler et al. \cite{rothenhausler2015backshift} extend LiNGAM to learn linear causal cyclic models in the presence of latent variables. The authors explore the setting where the equilibrium data of a model is observed which is characterized by a set of linear relations Eq.(\ref{LiNGAM}).

\begin{equation}C P(\mathbf{B}) ={\max _{{\left(m_{1}, \ldots, m_{\eta}, m_{\eta+1}\right)}_{1<\eta \leq p}\text { cycle }}} \prod_{1 \leq k \leq \eta}\left|\mathbf{B}_{m_{k+1}, m_{k}}\right|\end{equation}

where $B$ is the connectivity matrix. For their experiments, they assume that the data in an environment $j$ are equilibrium observations of the model

\begin{equation}\label{Backshift_Eqn}\mathbf{x}_{j}=\mathbf{B} \mathbf{x}_{j}+\mathbf{c}_{j}+\mathbf{e}_{j}\end{equation}

where $c_j$ is the random intervention shift. Given these locations, the interventions shift the variables by a value determined by $c_{j}$. $\mathbf{\Sigma}_{\mathbf{c}, j}$ is a diagonal matrix which is equivalent to demanding that interventions at different variables are uncorrelated. The final output of the model is the estimated connectivity matrix $\hat{B}$.

In \cite{huang2019causal}, authors propose a time-varying causal model to represent the underlying causal process in nonstationary time series. The authors assume that the time series in consideration are generated from the LiNGAM process.  They further allow each causal coefficient $b_{ij,t}$ and noise variance $\sigma_{i, t}^{2}$ to change over time which are modelled using the following autoregressive models:
\begin{equation}\begin{array}{l}
b_{i j, t}=\alpha_{i j, 0}+\sum_{p=1}^{p_{l}} \alpha_{i j, p} b_{i j, t-p}+\epsilon_{i j, t} \\
h_{i, t}=\beta_{i, 0}+\sum_{q=1}^{q_{l}} \beta_{i, q} h_{i, t-q}+\eta_{i, t}
\end{array}\end{equation}

where $\epsilon_{i j, t} \sim \mathcal{N}\left(0, w_{i j}\right), \eta_{i, t}$ $\mathcal{N}\left(0, v_{i}\right),$ and $h_{i, t}=\log \left(\sigma_{i, t}^{2}\right)$ models the volatility of the observed time series. The authors further state that, the time-varying linear causal model is actually a specific type of nonlinear state-space model with respect to hidden variables $b_{ij}$ and $h_{i}$. The causal graph is determined from the sampled particles. To determine whether there is a causal edge from $x_{j}$ to $x_{i}$ , the authors check both the mean and the variance of $\hat{b}_{i j, t}$.  Specifically, if both $\overline{\hat{b}}_{i j}=\frac{1}{T} \sum_{t=1}^{T} \hat{b}_{i j, t}<\alpha$ and $\frac{1}{T} \sum_{t=1}^{T}\left(\hat{b}_{i j, t}-\overline{\hat{b}}_{i j}\right)^{2}<\alpha,$ then there is no causal edge between $x_{j}$ and $x_{i}$.

Peters et al. \cite{peters2013causal} propose a class of structural equation models called Time Series Models with Independent Noise (TiMINo) for identifying the causal structure of a time series. The authors provide both theoretical analysis which provides general identifiability results and practical solution which introduces an algorithm based on testing the independence of the residuals for the case where there are no feedback loops between time series.

\subsubsection{Deep Learning based methods}
In this section we briefly review the causal discovery frameworks which utilize the power of deep neural networks to perform causal discovery on time series data and overcome the shortcomings of traditional time series causal discovery frameworks.

 Methods on Granger causality are built upon linearity of time series. However, in real world cases dependencies among time series are usually non linear and ignoring such interactions could lead to inconsistent estimation of Granger causality \cite{terasvirta2010modelling}. To incorporate non linear interactions into Granger causality detection, Tank et al. \cite{tank2018neural} propose a class of non linear architectures such as Multi Layer Perceptron (MLP) and Recurrent Neural Network (RNN) in which, each time series is modeled using an MLP or RNN. The input to the non linear framework is the past lags of all series and the output is the future value of a single series. The authors also leverage a group lasso penalty to further shrink the weights of the inputs to zero. 
 Dang et al. \cite{dang2018seq2graph} propose a deep learning based framework which consists of multiple customized gated recurrent units (GRUs) designed to discover nonlinear and inter-time series dependencies. More specifically, the paper introduces a new dual purpose recurrent neural network which models the lagged dependencies in each time series and leverages those to discover inter-timeseries dependencies. 
In \cite{wunonlinear}, the authors introduce a novel minimum predictive information regularization method to infer causal relations from time series, allowing deep learning models to discover nonlinear causal relations.  This work makes two assumptions regarding the causality. First is the causal sufficiency assumption which states that each time series $x^{(i)}$ can only be caused by the time series from $x ^{(1)}, x^{(2)}, ...x^{(N)}$. The second assumption is  the ``causality in mean" assumption which states that the causal relations influence the mean value of other variables. Their model tries to answer the question, how much can $X_{t-1}^{(j)}$ be corrupted without making the prediction of $x_{t}^{(i)}$ noticeably worse?
To do so, they take the input and add independent noise with learnable amplitudes and measure the extent of corruption by mutual information between input and corrupted output.

The risk for the mentioned situation can be given by:

\begin{equation}\begin{aligned}
R_{\mathbf{X}, x^{(i)}}\left[f_{\theta}, \boldsymbol{\eta}\right] &=\mathbb{E}_{\mathbf{X}_{t-1}, x_{t}^{(i)}, \boldsymbol{\epsilon}}\left[\left(x_{t}^{(i)}-f_{\theta}\left(\tilde{\mathbf{X}}_{t-1}^{(\boldsymbol{\eta})}\right)\right)^{2}\right] \\
&+\lambda \cdot \sum_{j=1}^{N} I\left(\tilde{X}_{t-1}^{(j)\left(\eta_{j}\right)} ; X_{t-1}^{(j)}\right)
\end{aligned}\end{equation}

where $\tilde{\mathbf{X}}_{t-1}^{(\boldsymbol{\eta})}:=\mathbf{X}_{t-1}+\boldsymbol{\eta} \odot \boldsymbol{\epsilon}\left(\right.$ or element-wise $,\tilde{X}_{t-1}^{(j)\left(\eta_{j}\right)}:=$, $\left.X_{t-1}^{(j)}+\eta_{j} \cdot \epsilon_{j}, j=1,2, \ldots N\right)$ is the noise-corrupted inputs with learnable noise amplitudes $\eta_{j} \in R^{K M},$ and $\epsilon_{j} \sim N(\mathbf{0}, \mathbf{I}) . \lambda>0$ is a positive hyperparameter for the mutual information $I(\cdot, \cdot) .$

At the minimization of $R_{\mathbf{X}, x^{(i)}}\left[f_{\theta}, \boldsymbol{\eta}\right]$, $W_{ji}$ is defined as $W_{ji}$ = $I\left(\tilde{X}_{t-1}^{(j)\left(\eta_{j}^{*}\right)} ; X_{t-1}^{(j)}\right)$, which is known as minimum predictive information and it measures the predictive strength of time series $j$ for predicting time series $i$, conditioned on all other observed time series.

Most existing methods in causal discovery rely on predefined kernels or data distributions. To relax such assumptions, Xu et al. \cite{xu2019scalable} propose a scalable causal discovery algorithm based on deep neural network. The proposed framework consists of four modules which account for temporal non linearity, learning the causal graph, identifying the intervariable non linearity and performing prediction for $X_{t}$ in the framework of Granger causality. In order to make the framework more scalable, the authors also propose to approximate the causal graph via k-rank matrix decomposition.

In \cite{pan2018hyperst}, the authors propose a hypernetwork framework for learning the intrinsic causality between spatial and temporal attributes to enhance the prediction performance of spatial temporal networks. They show that the spatial characteristics have a huge influence on the temporal characteristics and in order to capture this influence they use a hypernetwork.


 
 Most of the aforementioned approaches are based on causal sufficiency assumption and therefore do not consider unobserved confounders. In such cases, observed confounders can be taken into account by controlling for them using for instance conditional Granger test \cite{liao2010evaluating}.
 However, in most real-world scenarios we cannot expect to have all possible confounders measured. 
 Meng \cite{meng2019estimating} addresses the problem of unobserved confounder in non linear Granger causality based methods by approximating the distribution of unobserved confounders using Variational autoencoder. This distribution is sampled to get the estimated counfounders which are used in the Granger test.



\input{Evaluation/evaluation}

\section{Conclusion and Future Work}
\label{future}
In this paper, we provide a comprehensive survey of causal inference tasks for time series data. We first categorize traditional ways of modeling this type of data and discuss state-of-the-arts in each category. We then discuss two of the most important causal inference tasks on time series, i.e. causal treatment effect estimation and causal discovery. Each of these tasks is classified based on the type of approaches proposed to solve the problem and for each category, state-of-the-arts are discussed. We also provide an extensive list of datasets and evaluation metrics used to assess the performance of frameworks proposed for different tasks. These metrics and datasets can be used as a guideline for future research in this field.
Lastly, in the following, we discuss some future research opportunities in terms of estimating treatment effects, causal discovery and evaluating the performance of the models for each task.

\textbf{Causal Treatment Effect Estimation:} 
Most of the methods in this task are base on the Stable Unit Treatment Values Assumption (SUTVA) which is that the potential outcome is not affected by exposure to the treatment of other units and there is no hidden variations of treatment. However, in many fields of research such as social science, friends, families and acquaintances have influence on subjects' awareness of treatment and their desire to follow it~\cite{jackson2020adjusting}. Hence, there is a need to provide methods that account for social network and peer-influence both in participation decisions and in determining
a subject’s outcomes. Moreover, existing methods usually model treatments as discrete events. However, some treatments, for example dialysis or intravenous diuretics \cite{soleimani2017treatment}, are carried out continuously over a period of time. Therefore designing methods for estimating the effect of continuous-time and continuous-valued treatments is a direction that needs to be explored.

\textbf{Causal Discovery:} Most exiting methods in causal discovery, especially methods based on deep neural networks, rely on the concept of Granger causality. Pearl's causality is another concept of causality which has been widely popular in i.i.d data scenarios \cite{ellis2008learning, heckerman2013bayesian, meganck2006learning}. One promising direction to pursue in future research on causal discovery for time series data is to utilize the power of deep neural networks to learn the causal structure from time series data. Moreover, most exiting algorithms in time series field only leverage observational data. Interventional data has proven to be extremely useful in learning more accurate causal structures from i.i.d data \cite{cooper2013causal, steyvers2003inferring, brouillard2020differentiable}. We therefore suggest leveraging both observational and interventional time series data to learn a better and more accurate causal structure .

\textbf{Performance Evaluation and Benchmark Datasets:} In terms of data, both treatment effect estimation and causal discovery tasks need to include more robust datasets for evaluation purposes. For example, in causal discovery, we need datasets to evaluate multi modal causal discovery algorithms when one of the modalities is time. Multimodal data, have gained a lot of attention in various fields like healthcare \cite{anwar2014multi} and financial \cite{lee2019multimodal}, thus promoting the need for such datasets for evaluation. Multimodal data represents data of different types like images, text, etc. For example, satellite images of a scene taken over different time represents multimodal data.
When it comes to treatment effect estimation, there is a need for datasets that are suitable for calculating the counterfactual outcomes along with the factual outcomes as they help in estimating the individual treatment effects which would help personalize the effect of a treatment on each individual rather than considering an average population. Most of the datasets mentioned in the earlier section are generated from observational studies and its impossible to obtain both factual and counterfactual outcomes in such a setting. There have been works like \cite{jaber2020causal, louizos2017causal}, that use constructed data which is a mix of observational and randomized control trials (RCT) to overcome this difficulty. We need more such data for better evaluation of treatment effect estimation.

\printbibliography


\end{document}

%% file: intro.tex
Time series data consist of ordered sequences of real-valued data which are often collected over time. With the rapid growth of time series data  generated by different domains such as bioinformatics, medical, neuroscience and financial applications, various approaches have been developed.
Research on time series data has been going on for over a decade and researchers have come up with different approaches to analyze this type of data for different purposes such as classification \cite{lines2014ensembles, roberts2002bayesian}, clustering \cite{li2011time,NEURIPS2019_1359aa93}, forecasting \cite{esm2002, wang2019deep}, estimating the impact of an intervention/treatment over time \cite{moodie2007demystifying, athey2017state} and discovering the causal relations between the time series components \cite{granger1969investigating, entner2010causal}.
In this survey, we focus on the last two tasks, i.e., estimating the effect of an intervention/treatment and identifying the causal relations and refer to them as causal inference for time series analysis.

Questions such as ``Was an enforced policy effective?" or ``Which medicine works better for a specific disease?" are crucial questions in law-making and medical fields and answers to these questions can help making important decisions/policies. In order to answer such questions with data, one needs to estimate the effect of an intervention/treatment. For example, to evaluate the effectiveness of tabacco control program, Abadie et al.  \cite{abadie2010synthetic} propose a framework to estimate the effect of this program on cigarette sales. In another example, Bica et al. \cite{bica2019time} propose a framework to predict patient's response to a specific medicine over time.
The task of estimating such effects from the data is called causal treatment effect estimation. Causal treatment effect estimation is one the most important tasks in causal inference which leverages concepts from causality to estimate this effect from the data.
The state-of-the-art causal treatment effect estimation approaches for time series data can be categorized  into three main types: 1)time-invariant treatment effect; 2)time-varying treatment effect; and 3)dynamic regimes. We discuss some widely used methods in each category.

Another principal task discussed in this survey is causal discovery. Causal discovery is the task of identifying the causal relationships between variables in the data. Causal discovery for time series data refers to the task of understanding and identifying interdependencies amongst individual components of a time series. This task is seen in a variety of applications such as economy and earth system science. For instance, causal discovery can be used to identify the performance indicators of stock analysis \cite{ hiemstra1994testing} or discover the causal relations between the external drivers of climate change and climate variables \cite{stips2016causal,runge2019inferring}. We classify causal discovery approaches for time series data into three main categories, namely, Granger causality and conditional independence based, structural equation model based, and deep learning based methods and discuss them in detail.

Despite having extensive surveys on non-causal time series analysis from different perspectives such as time series classification \cite{abanda2019review}, deep learning and unsupervised feature learning \cite{fawaz2019deep} and data mining approaches \cite{FU2011164}, no existing survey reviews the current progress of algorithms which are designed to analyze time series data from a causal perspective.  
Different from existing efforts, in this paper, we discuss the state-of-the-art methods for causal inference for time series analysis and its two main tasks. Since the evaluation of causal inference in general and causal inference on time series in particular is a challenging task, we also enlist some benchmark datasets and evaluation metrics which are commonly used by the researchers.  

We first explain prevalent methods and concepts used for modeling time series data (Section \ref{modeling}). We then discuss necessary definitions and assumptions for causal inference on time series data  which are used in the rest of the paper (Section \ref{sec:causal_background}). Next, we discuss causal treatment effect estimation and causal discovery for time series (Sections \ref{causaltreatment} and \ref{discovery}, respectively). We then provide extensive guidelines on how the frameworks proposed for each of these tasks can be evaluated by presenting a list of commonly used datasets and evaluation metrics (Section \ref{eval}). We conclude this work with some future directions (Section \ref{future}).




%% file: TimeSeriesModelling/timeSeriesModelingNew.tex
\section{Modeling Time series data}
\label{modeling}
Time series data is a sequence of real-valued data with each data point related to a timestamp. Mathematically, time series data is denoted as X(t) = ({$x_{1}(t)$, $x_{2}(t)$,..., $x_{k}(t)$}) where k is the number of variables measured at a discrete timestep t $\in  \mathbb{Z}$. In this section we discuss different techniques to model a time series data.

\subsection{Autoregressive Models}

   One of the earlier methods to model time-series data is the AutoRegressive Integrated Moving Average (ARIMA) model. The ARIMA models the time-series assuming three fundamental relationships between the time series - autoregressive, moving average, and differencing. The Autoregressive (AR) component determines the value of a current timestamp $X(t)$ from a finite set of previous timestamp values of some length p and some error $\epsilon$. The order of autoregression is the number of preceding timestamps used to determine the value of the current timestamp, given by 
   \begin{equation}
       AR(p) =  \sum_{i=1}^{p} a_{i}.X(t-i) + c + \epsilon_{t},
   \end{equation}
   
   where $a_{i}$ and p are the coefficients and the order of the AR model, respectively.
   The moving average (MA) component models the value of a current timestamp $X(t)$ as a linear combination of the prediction errors($\epsilon_{t}$) at the previous timestamps of length q, where q is the order of the moving average component.
   
   \begin{equation}
       MA(q) =  \sum_{i=1}^{q} b_{i}.\epsilon_{t-i} + \mu + \epsilon_{t},
   \end{equation}
   
   where $b_{i}$, $\mu$ and q are the coefficients, mean of the series and the order of the MA model, respectively. 
    
   The AR and MA components are enough to model a time series in the case the time series data is stationary i.e. the time series have the same values of specific properties (mean, variance) over every time interval. In the case of non-stationary data, as shown in Equation \ref{eq:nonToStationary}, the time series data is differenced with a shifted version of itself to make the data stationary.
   
   \begin{equation}
   \label{eq:nonToStationary}
       Y(t) = X(t) - X(t-r),
   \end{equation}
   
   where r is the order of differencing.
   
   The primary task in this type of modeling is estimating the coefficients $a_i$ in AR and $b_{i}$ in MA models, as well as the orders p, q, and r of the AR, MA, and differencing, respectively.  The Box and Jenkins~\cite{boxJenkins1968} approach is used to estimate the parameters of an ARIMA model, assuming an  underlying  model,  and  verifying  if  the  residuals  or  error term is a random distribution.  This process is repeated with different models until the right model is obtained. The Simple Exponential Smoothing (SES)~\cite{brown1956exponential} is a modification of the ARIMA model with exponential weights assigned to each observation. Double and Triple Exponential Smoothing (DES and TES)~\cite{esm2002} handles non-stationary data by introducing an additional parameter $\beta$ for smoothing the trend in the series. Moreover, an extra  parameter $\gamma$ is  introduced  in  TES  to  control  the influence of seasonality.
   
\subsection{Dynamic Bayesian Networks}

Dynamic Bayesian Networks (DBNs) are an extension of Bayesian Networks (BNs) to model the evolution of random variables as a function of a discrete timestep sequence, represented as a directed acyclic graph. Formally, a Bayesian Network is defined by $G = (V, E$), where $V$ and $E$ are the set of nodes and edges. The conditional probability distribution of the set of nodes $V$ can be expressed as the factorized joint probability given by:

\begin{equation}
    P(V) = \prod_{x \in V} P(x|\pi_{x}),
\end{equation}

where $\pi_{x}$ are the parents of node $x$. A DBN is represented as a pair of two Bayesian networks $B_{p}$ and $B_{2d}$. $B_{p}$ is a BN modeling the prior distribution of the random variables at time 1. $B_{2d}$ is a two slice BN representing the transition from time t-1 to time t, as a probability distribution P($x_{t}|x_{t-1})$ for nodes x belonging to V by means of a directed acyclic graph G =(V,E) as follows: 

\begin{equation}
    P(V_{t}|V_{t-1}) = \prod_{x \in V, \pi_{x} \in V} P(x_{t}| \pi_{x_{t}})
\end{equation}

If we define T as the total length of the path, the joint distribution of the sequence is given by: 

\begin{equation}
    P(V_{0:T}) = \prod_{x \in V} P_{B_{p}}(x_{1}|\pi_{x_{1}}) \times \prod_{t=2}^{T} \prod_{x \in V} P_{B_{2d}}(x_{t}|\pi(x_{t})) 
\end{equation}

Typically, the variables in a DBN, are partitioned into two sets of variables, $V_{t} = (Z_{t}, X_{t})$, representing the hidden and output (observed) variables of a state-space model.

The parameters of the DBN can be learned from the data. Based on the probability distributions and the assumptions made on the dynamics (in the case of observable data), Maximum Likelihood Estimation (MLE) or Maximum A Priori (MAP) is used. For hidden variable models, the parameters are generally learned using the Expectation-Maximization (EM) algorithm. 

Next, we discuss two of the state-of-the-art DBN models used for time series modeling. 

\subsubsection{State-Space Models}

 State-space models use a latent state $z_{t} $ to model the time series data, i.e. encoding time series components level, trend and seasonality patterns. An SSM is denoted by a state-transition equation, which describes the transition dynamics $p(z_{t}|z_{t-1})$ of the evolution of the latent state over time. It also represents an observation model that describes the conditional probability $p(x_{t}|z_{t})$ of observations given the latent state.
  A widely used example of SSM is the linear dynamical system (LDS), where the states are real-valued and change linearly with time, satisfying the first-order Markov assumption. The LDS can be expressed by the following equations:
  \begin{align}
       z_{t} = Az_{t-1} + \eta , z_{t} \in \mathbb{R}^{k}, \eta \sim N(0, Q) \\
      x_{t} = Cz_{t}  + \epsilon, x_{t} \in \mathbb{R}^{d}, \epsilon \sim N(0, R),
  \end{align}
  
 where $A \in\mathbb{R}^{k X k}$, $C \in \mathbb{R}^{d X k}$, while Q and R are covariance matrices. The joint probability for the states and observations is given by: 
 \begin{equation}
     P(z_{t}, x_{t}) = P(z_{1})P(x_{1}|z_{1})\prod_{t=2}^{T}P(z_{t}|z_{t-1})P(x_{t}|z_{t})
 \end{equation}
 The Kalman filtering algorithm~\cite{kalman1960} is used to perform inference tasks such as filtering ($p(z_{t}|x_{1:t})$), smoothing ($p(z_{t}|x_{1:T})$), and prediction ($p(x_{t}|x_{1:t-1})$). 
 
 Nonlinear extensions of the LDS is proposed in unscented Kalman Filter~\cite{Julier97anew}, by generalizing the state transition and emission to nonlinear functions. Zheng et al.~\cite{zheng2017state} proposed State-space LSTM which modeled the sequential latent states by parameterizing the transition function between states by a neural network. The Particle Gibbs~\cite{Andrieu10particlemarkov} is used for parameter estimation of the model which samples from the joint posterior, eliminating the need to sample at each time point thus removing the assumption of a factorizable posterior.  Karl et al.~\cite{karl2017deep} uses the stochastic gradient variational Bayes to learn the latent state dynamics under a nonlinear Markovian setting. Instead of a deterministic f and g functions, Wang et al.~\cite{Wang06gaussianprocess} placed Gaussian process priors over both the nonlinear functions f and g and found a MAP estimate of the latent variables. 
 
 Switching state-space models~\cite{Ghahramani96switchingstate-space}, model the observations $x_{t}$ using M real-valued hidden state-space vectors $z_{t}^{m}$ and one discrete state vector $s_{t}$. A multinomial variable with M possible values represents the discrete state variable $s_{t} \in ({1, ..., M}) $ which acts as a switch variable. The observed variable is represented using the state-space model $m$, conditioned on this discrete state. The discrete state follows a Markovian dynamics with a specified initial state($p(s_{1})$) and transition probability matrix($p_{s_{t}|s_{t-1}}$). The real-valued state variables have linear-Gaussian dynamics with each variable having its transition matrix, initial state, and noise. The joint distribution of the observed and hidden variables is given by:
 
 \begin{equation}
   \begin{split}
    P(s_{t}, z_{t}^{1}, ... , z_{t}^{m}, x_{t}) & = P(s_{1})\prod_{t=2}^{T}P(s_{t}|s_{t-1}) \\ & \times \prod_{m=1}^{M} P(z_{1}^{m}) \prod_{t=2}^{T} P(z_{t}^{m}|z_{t-1}^{m})  \\ 
    & \times \prod_{t=1}^{T} P(x_{t}|z_{t}^{1},...,z_{t}^{M}, s_{t})
   \end{split}
 \end{equation}
 
 The discrete switch variable acts as a gating network for the M real-valued states. Gibbs sampling is used to approximate the marginal probabilities, required for the evaluation of the expectations for learning the model parameters using the EM algorithm. 
 
 \subsubsection{Hidden Markov Model}
 
 The Hidden Markov Model (HMM) is another class of SSM, where the states are assumed to be discrete and distributed according to the Markov Process. The joint probability distribution of the discrete hidden state $s_{t}$ and the observed sequence $x_{t}$ can be denoted similarly to the SSM as: 
 
 \begin{equation}
     P(s_{t}, x_{t}) = P(s_{1})P(x_{1}|s_{1}) \prod_{t=2}^{T} P(s_{t}|s_{t-1})P(x_{t}|s_{t})
 \end{equation}
 
 The Baum-Welch algorithm~\cite{baum1970}, a type of EM algorithms for HMM, is used to learn the parameters of this model. 
 
In many applications, the HMM model is extended by representing the $P(x_{t}|s_{t})$ using a mixture of Gaussians for each state ($M_{t}$), forcing $x_{t}$ to get the information from $s_{t}$ bottlenecked through $M_{t}$.

Autoregressive HMMs~\cite{shannonWilliam2011} relaxes the HMM assumption of conditional independence of observations given the hidden state, by allowing $x_{t}$ to be connected to $x_{t-1}$ along with the hidden state $s_{t}$. 

Factorial HMMs~\cite{Ghahramani96factorialhidden} extend the HMM by having a collection of discrete state variables, as opposed to a single state variable for the original HMM. The state variable $s_{t}$ is represented as a combination of ($s_{t}^{1}$, ... , $s_{t}^{M}$), each of which can take K possible values. This will extend the state space to $K^{M}$ possible values, which is equivalent to a regular HMM with $K^{M}$ states. Ghahramani et al.~\cite{Ghahramani98learningdynamic} proposes a constraint on the interactions of the state variables, in which the state variables evolve through the following dynamics: 

\begin{equation}
    P(s_{t}|s_{t-1}) = \prod_{m=1}^{M} P(s_{t}^{m}|s_{t}^{m-1}),
\end{equation}

thus uncoupling the different state variables from each other. 

The uncoupling of state variables was relaxed in Saul et al.~\cite{Saul98mixedmemory} by coupling the variables of a time step in order, i.e $s_{t}^{m}$ interacts with $s_{t}^{n}$ for 1 $\leq$ n $<$ m. The model parameters are learned using the EM algorithm, where the marginal probabilities used for the expectations are approximated using the Gibbs sampling algorithm. 
 
 \subsection{Gaussian Processes}


Taking advantage of available data and performing robust analysis is only practical through modeling the uncertainty. Therefore, the Bayesian inference is leveraged to handle uncertainty in a noisy and dynamic environment. Gaussian Processes are a class of Bayesian nonparametric models that are particularly suitable for modeling time series data. In particular, Gaussian Processes (GPs) are a class of stochastic processes, which define a joint Gaussian distribution over a collection of random variables. A function ($f(x)$) which follows a Gaussian process, is specified by the mean (m(x)) and covariance ($k(x,x^{'}$) functions, denoted as $f(x) \sim GP(m(x), k(x,x^{'}))$. Formally, the gaussian process responsible for generating $Y$ given $X$ is given by,
\begin{equation}
\mathbf{y}_{n}=f\left(\mathbf{x}_{n}\right)+\epsilon_{n}, \quad \epsilon \sim \mathcal{N}\left(\mathbf{0}, \sigma_{\epsilon}^{2} \mathbf{I}\right)
\end{equation}

where $\epsilon$ is the Gaussian noise term.
In what follows, we briefly discuss the Gaussian processes frameworks.

\subsubsection{Deep Learning and GP} 
Motivated by the success of Deep models in different tasks, recently, several attempts have been made to combine deep models with Gaussian processes~\cite{damianou2013deep, wilson2016stochastic, wilson2016deep, maddix2018deep} and create Deep Gaussian Process (Deep GP) models. These frameworks typically use neural networks to map the input to the feature space (extract non-stationary features) whereas the last layer sparse Gaussian process performs regression over the latent space.
For example, Wilson et al.~\cite{wilson2016deep} propose to leverage fully connected and convolutional neural networks as input to the spectral mixture base kernel and use local kernel interpolation~\cite{wilson2015kernel}, spectral mixture covariance functions~\cite{wilson2013gaussian}, inducing points~\cite{quinonero2005unifying} and structure exploiting algebra~\cite{saatcci2012scalable} and create more powerful and expressive closed-form covariance kernel for Gaussian Processes.
In another attempt, Maddix et al.~\cite{maddix2018deep} propose a scalable hybrid model which combines both deep neural network and classic time series model to perform accurate forecasting which also takes uncertainty into account.The model consists of a global deep neural network and a local Gaussian Process model.

\subsubsection{GP Methods for Inference} 
Havasi et al.~\cite{havasi2018inference}  propose an inference method for Deep Gaussian Processes models based on the Stochastic Gradient Hamiltonian Monte Carlo method. The paper also shows that the posterior in these models is of non-Gaussian nature and therefore, the existing approaches based on the variational inference that estimate a Gaussian posterior are poor potential approximations for the multimodal posterior.\\
The authors in~\cite{li2016scalable} propose an uncertainty-aware classification framework that facilitates learning black-box classification models for classifying sparse and irregularly sampled time series. The framework uses Gaussian Process regression to transform the irregular time series data into a uniform representation, which permits sparse and irregularly sampled data to be fed into any black-box classifier which is learnable using gradient descent while preserving uncertainty.
Tobar et al.~\cite{tobar2015learning} propose a framework called Gaussian Process Convolutional Model (GPCM), which serves as a generative model for stationary time series. The main idea behind this model is based on the convolution between a filter function and a white noise process. This approach recovers a posterior distribution over the spectral density directly from the time series. It also places the nonparametric prior over the spectral density before recovering the posterior distribution. Learning the model from the data allows performing inference on the covariance kernel as well as the spectrum in a probabilistic, analytic, and computationally tractable manner.
Cunningham et al.~\cite{cunningham2012gaussian} propose a Gaussian Process model for analyzing multiple time series with multiple time markings. The proposed model can be considered as a mapping between the input space and the given data markers. Because of this, the model can be used as a choice for a covariance function. It also facilitates learning and inference to be standard.
HajiGhassemi et al.~\cite{hajighassemi2014analytic} propose an algorithm for long term forecasting with periodic Gaussian processes. They also state that, for long term forecasting it is necessary to map the probability distributions through Gaussian Processes. They use re-parameterization of a commonly used stationary periodic kernel which in turn allows them to employ an analytic double approximation strategy to compute the moments of the predictive distribution.\\
For a comprehensive surveys on Gaussian Processes and how they can be used to model time series data and scaled to big data scenario, we refer the reader to three studies provided by Rasmussen~\cite{rasmussen2003gaussian}, Roberts et al.~\cite{roberts2013gaussian}, and Liu et al.~\cite{liu2020gaussian}.



\subsection{Neural Networks}

    For more complex, noisy, and higher dimensional real-world data model assumption techniques such as ARIMA, state-space models is not an efficient technique since the dynamics are unknown or too complex~\cite{Taylor2009ComposableDM}. Various unsupervised generative models have been extended for time series data to solve this problem. Graves et al.~\cite{gravesRNN2013} proposed a method for sequence generation using recurrent models such as RNN and LSTM, by processing the real data at each step ($x_{t}$) and predicting the value of the next step ($x_{t+1}$). Output predictions ($y_{t}$) at each step are made probabilistic and sampled from to be fed as the next step input. Iteratively sampling at each step from the already trained network, and passing it to the next step produces a  novel sequence. Although theoretically possible, RNNs in practice suffer to capture long-term dependencies, and thus LSTMs are used which are shown to capture long term dependencies with the help of various gating mechanisms. Additionally, the hidden layers are stacked to increase the depth across space to allow for higher non-linearities to be captured at each time step. The probability of the input sequence x is given by Pr(x) = $\prod_{t=1}^{T}Pr(x_{t+1}|y_{t})$. 
    
    Restricted Boltzmann Machine(RBM), a generative probabilistic model between the input nodes (observable) and latent nodes (hidden), connected by a weight matrix (W) and having associated bias vectors c and b respectively, is extended for sequential data in various works. The model is generally trained by minimizing the reconstruction error using contrastive divergence~\cite{Hinton2002a}. Conditional RBM and temporal RBM~\cite{SutskeverH07} extend the RBM model with a connection between the current hidden units and the past observable units along with auto-regressive weights for capturing short-term temporal patterns. The dependency between the bias vectors and the past visible units are defined by, 
    
    \begin{equation}
    \begin{aligned}
        b_{j}^{'} = b_{j} + \sum_{i}^{n} B_{i} x(t-i)\\
        c_{i}^{'} = c_{i} + \sum_{i}^{n} A_{i} x(t-i),
    \end{aligned}
    \end{equation}
    
    where $B_{i}$ and $A_{i}$ are the weight matrices connecting respectively current hidden units and current observable units to observable units at time $t-i$. Thus the conditional probabilities for activation of the hidden units and the visible units become: 
    
    \begin{equation}
    \begin{aligned}
        P(h_{j}|x) = \sigma(b_{j} + \sum_{i} W_{ij}x_{i}) + \sum_{k}\sum_{i} B_{ijk} x_{i}(t-k)\\
        P(x_{i}|h) = \sigma(c_{i} + \sum_{i} W_{ij}h_{j}) + \sum_{k}\sum_{i} A_{ijk} x_{i}(t-k)
    \end{aligned}
    \end{equation}

   Oord et al.~\cite{oord2016} proposed a CNN based architecture called WaveNet to generate audio waveforms. WaveNet tries to approximate the joint probability of the time series $X = ({x_{1}, x_{2}, ..., x_{T}}$) by making $x$ depend on all the previous samples. 
   WaveNet uses a special convolutional layer called the dilated causal convolution. A causal convolution forces the prediction at any timestep to depend on the previous timesteps and  prevents dependencies on future timesteps. Dilated convolutions is a filter which can span larger than it's length by dilation with zeros, skipping input values at some steps, thus increasing the receptive field of the filters with increasing depth. Additionally, gated activated units are used which allows the network the ability to preserve and forget certain input values. Undecimated Fully Convolutional Network (UFCNN)~\cite{mittelman2015}, uses Fully Convolutional Layers with 1-D causal filters, and the filter's at $l^{th}$ resolution level is upsampled by a factor of $2^{l-1}$ along the time dimension along with removal of max-pooling layers and other upsampling operators.
   
   Generative Adversarial Network (GAN) based models have also been proposed for Time Series Generation. Mogren et al.~\cite{mogrenCRNN2016} proposed C-RNN-GAN to generate continuous sequential data by modeling the joint probability distribution of the sequence. The generator was designed using an LSTM and the discriminator consisted of a bidirectional RNN. The model was trained using the standard GAN loss. Yoon et al.~\cite{yoonTSGAN2019} argued that using recurrent networks for the generator and discriminator, and summing the GAN loss over sequences is not enough for capturing the temporal dynamics of the data. They proposed a stepwise supervised loss along with the unsupervised adversarial loss to encourage the model to capture the stepwise temporal dependencies.

%% file: CausalBackground/Background.tex
\section{Causal Inference}
\label{sec:causal_background}

In this section, we briefly review the concepts from causal inference used in this paper for causality based time series models. 
Below are the common methods  in calculating the difference between the outcome of the data  with intervention and the control group.

Suppose $A$ is a treatment dummy random variable and $Y$ is the desired outcome random variable, then $A$ can
take value $a=1$ or $a=0$ denoting the presence or absence of a treatment, respectively. In this case, $Y_a$ will be defined as the potential outcome under exposure to the treatment (i.e. $Y_{a=1}$) or under control (i.e. $Y_{a=0}$). $\delta Y_a = Y_{a=1}-Y_{a=0}$ denote the individual (or unit) causal effect (ITE) of the treatment.

\begin{definition}[Average Treatment Effect]
The average causal effect also known as the average treatment effect of the population can be calculated as follows:
    \begin{equation}
                ATE=\mathbb{E}[Y_{a=1}-Y_{a=0}], 
    \end{equation}
and it is non-zero when the treatment A has a causal effect on the mean of the outcome.
\end{definition}
 This measure implicitly assumes that the individuals are drawn from a large population. However, due to selection bias, the units might not be representative of such a population. In this case, sample average treatment effect (SATE) will be used that only calculates the treatment effect of the units in that specific study, which avoids any assumption on the distribution of the samples \cite{balzer2015targeted},
 \begin{definition}[Sample Average Treatment Effect]
\begin{equation}
            SATE=\sum_{i=1}^{m}[Y_{i,a=1}-Y_{i, a=0}], 
\end{equation}
where m is the number of samples and $Y_{i,a}$ is the outcome of sample $i$ under the treatment $a$.
 \end{definition}
On one hand, individual treatment effects in the population might be heterogeneous, meaning that the treatment affects the individuals or sub-populations differently. In this case, it is more desirable to consider the conditional average treatment effect (CATE) \cite{abrevaya2015estimating} to calculate the effect of a treatment on the sub-population.


\begin{definition}[ Conditional Average
Treatment Effect]

In the randomized controlled trial, the Conditional Average Treatment Effect (CATE) is estimated as follows:
\begin{equation}
\begin{aligned}
            CATE=\mathbb{E}[Y_{a=1}-Y_{a=0}|X=x]
\end{aligned}
\end{equation}
where $X$ is the covariates (or features) and $x$ is the values that the covariates take.
\end{definition}

On the other hand, we might be interested in the causal treatment effects for only those individuals of the population who choose to participate in the treatment. In this case, we can calculate the average treatment effect of the treated (ATT) sub-population as follows:
\begin{definition}
[Average Treatment Effect of the Treated]
\end{definition}
\begin{equation}
\begin{aligned}
            ATT=\mathbb{E}[Y_{a=1}-Y_{a=0}|A=1]
\end{aligned}
\end{equation}

All the above estimators are true if we conduct experiments on randomized trials. However, researchers may only have access to outcome values reported at the aggregate level (observational data). To obtain consistent estimators from observational data, identifiability conditions should be hold \cite{rosenbaum1983central, hedeker2006longitudinal, hernan2010causal}:

\begin{assumption}[Consistency] This assumption indicates that if $Y_{a=1}$  denotes the potential outcome for the treated subject then its value is known and is equal to the the observed outcome, $Y$. Although $Y_{a=0}$ (potential outcome under control) remains unknown. This is also true for the untreated subject. In other words, for the treatment variable $A$, if $A=a$, then $Y_a=Y$.

\end{assumption}
\begin{assumption}[Positivity]
The probability of receiving every value of treatment conditional on some measured covariates $X$ is 
greater than zero. In other words, $Pr[A=a|X=x]>0$ for all values x with $Pr[X=x]\neq 0$, in the population of interest.
\end{assumption}

\begin{assumption}[Conditional Exchangeability] Un-conditional exchangebility implies that the treatment group, had they been untreated, would have experienced the same distribution of outcomes as the control group. While in conditional exchangebility, the conditional probability of receiving every value of treatment, depends only on measured covariates $X$, i.e. $Y_a$ and $A$ are statistically independent given every possible value for X.

\end{assumption}

Below we list some common assumptions required to perform causal discovery in observational data.

\begin{assumption}[Causal Stationarity]
The time series process X with graph defined over it is called causally stationary over a time index set T if and
only if for all links $x^i_{t-T}$
\begin{equation}
            x^i_{t-T} \notindependent. x^j_{t} | X^-_{t} \backslash {x^i_{t-T}} \text{ holds for all } t \in T
\end{equation}

\end{assumption}

\begin{assumption}[Causal Sufficiency]
A set of variables is causally sufficient for a process, if and only if it includes all common causes of every two pairs in the set.
\end{assumption}

\begin{assumption}[Causal Markov Condition]
The joint distribution of a time series process X with
graph G fulfills the Causal Markov Condition if and only if
for all $Y_t \in X_t$ with parents $P_{Y_t}$ in the graph:
\begin{equation}
            X_{t}^{-} \backslash P_{Y_t} \text{~d-separated~} Y_t | P_{Y_t} \implies X_{t}^{-} \backslash P_{Y_t} \indep Y_t | P_{Y_t},
\end{equation}

\end{assumption}

\begin{assumption}[Faithfulness]
The joint distribution of a time series process X with
graph G fulfills the Faithfulness condition if and only if for
all disjoint subsets of nodes (or single nodes) $A, B, S \subset G$ it holds that
\begin{equation}
            X_A  \indep  X_B | X_S \implies A \text{~d-separated~} B | S
\end{equation}
\end{assumption}

\section{Causality and Time series Analysis}
In this section we discuss two most important tasks in causal inference for time series data, i.e. causal treatment effect estimation and causal discovery. We first explain causal treatment effect estimation problem, classify the existing approaches based on different settings given the time and explain the state-of-the-art approaches in each category. We then discuss the task of causal discovery for time series data, categorize the proposed frameworks based on the type of model being used and explain them in detail.



%% file: CTEEstimation/CTE-Intro.tex
\subsection{Causal Treatment Effect Estimation on Time Series}
\label{causaltreatment}
Policymakers often face challenges to assess the impact of an intervention (i.e., a change in policy) on an outcome of interest. For example, a state government wants to estimate the effect of a tobacco control program on cigarette sales using the available data before and after a proposition~\cite{abadie2010synthetic}.

There is a need to evaluate both the positive and negative valued consequences of the designed or unintended policies and interventions to ascertain whether they were effective or not.

To this end, researchers proposed various methodologies that account for different settings based on the time to estimate the treatment effect: (1) time-invariant treatment effect, (2) time-varying treatment effect, and (3) dynamic regimes. 

In this section, we will introduce the recent developments and existing applications of causal treatment effect estimation based on the aforementioned categories specifically designed for time series data.

\input{CTEEstimation/Time-invariant}
\input{CTEEstimation/Time-varying}
\input{CTEEstimation/DynamicRegimes}

%% file: CTEEstimation/Time-invariant.tex
\subsubsection{Time-invariant Treatment Effect}

A treatment is time-invariant or fixed when it occurs at one specific point of time and then does not change afterward, for example, a one-dose drug.
More formally, let $X(t)$ be the time series outcome recorded at times $t =1, 2, ..., n$, and let $A$ be a dichotomous treatment that can take values $a=0$ (untreated) or $a=1$ (treated). We will have $X_a$ that denotes the potential outcome for the treatment group (when $a=1$) or control group (when $a=0$). At $1 \leq T \leq n$ an intervention occurs. We are mostly seeking to model the potential outcome of the treatment group at $t>T$ had it not received the treatment ($X_{a=1}^c(t>T)$) which is known as the counterfactual outcome. The counterfactual tells us what would have happened in the treatment group had we not applied the policy. The difference between the observed values for the treatment group ($X_{a=1}(t>T)$) and the counterfactual outcome (unobserved values) would give us an estimation for the treatment effect. This difference can be reported as a numerical value using measures introduced in section \ref{sec:causal_background} such as ITE, ATE, and ATT. In the following section, we will introduce common time-invariant causal effect estimation methods. 

\begin{figure}[ht]
  \centering
  \includegraphics[width=1\linewidth]{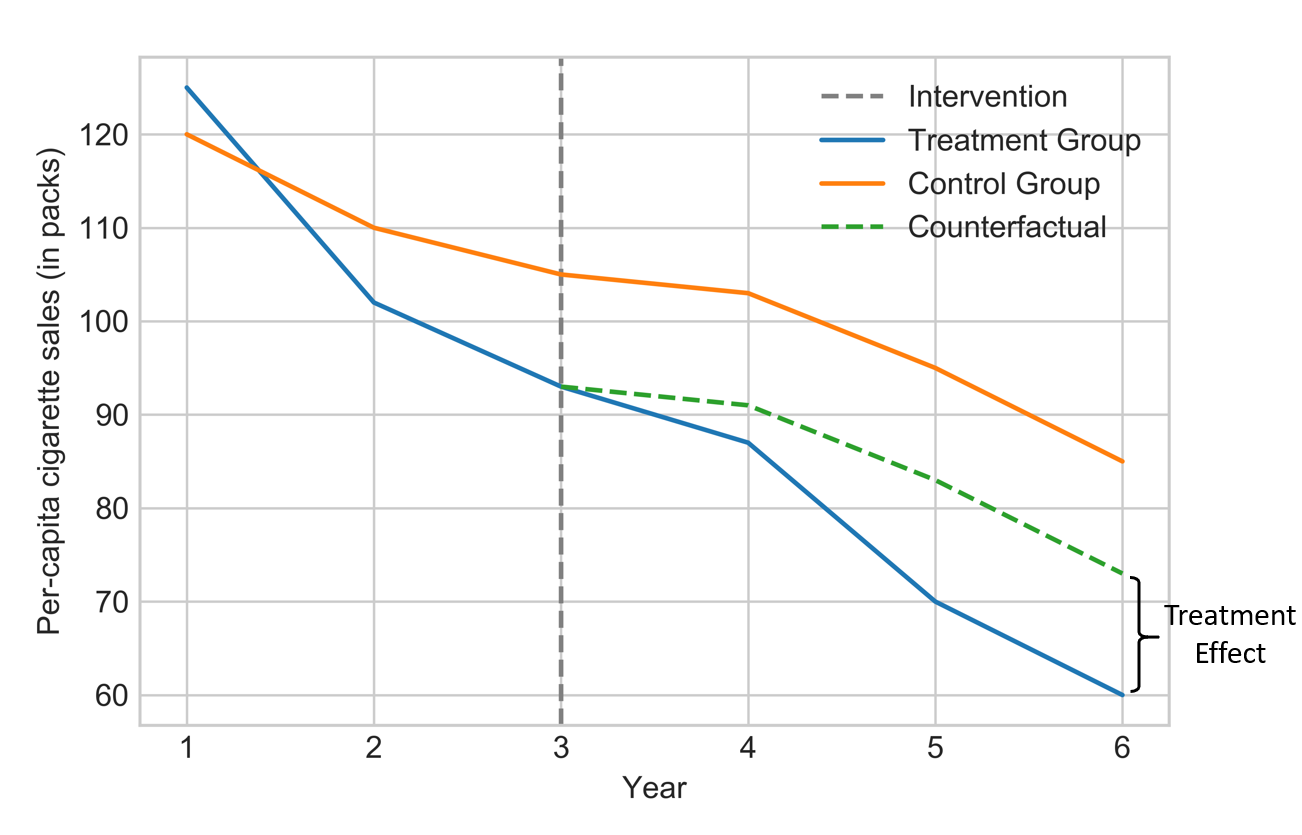}
  \caption{An illustration of the DiD method on a hypothetical example inspired by \cite{abadie2010synthetic}. The treatment group receives the intervention (e.g. tobacco control program) on the third year. The counterfactual (i.e. the green dotted line) tells us what would have happened in the treatment group had everything evolved as it did in the control group. The difference between the green dotted line and the blue line after the intervention is the treatment effect. The result shows that the tobacco consumption fell markedly following the tobacco control program.} 
  \label{fig::diff-in-diff} 
\end{figure}

An important tool mostly used by the econometricians to capture the causal effect of the time series data before and after the treatment is \textit{difference-in-differences} (DiD) \cite{athey2017state}. This tool goes by an assumption called \textit{common trends} or \textit{parallel trends} \cite{angrist2014mastering} that uses the change of the outcome of the control group as a counterfactual for the treatment group in the absence of the treatment. Figure \ref{fig::diff-in-diff} illustrates a hypothetical example of DiD on cigarette sales before and after the tobacco control program. The aim is to see whether this program affected per-capita cigarette sales.

If $t<T$ and $t>T$ denotes the  pre- and post- treatment periods (e.g. $T=3$ in Figure \ref{fig::diff-in-diff}), respectively, then we can calculate the DiD measure using the ATT metric as follows \cite{athey2006identification}:
\begin{equation}
\begin{aligned}
            DiD = \{\mathbb{E}[X(t>T)|A=1]-\mathbb{E}[X(t<T)|A=1]\}\\ -\{\mathbb{E}[X(t>T)|A=0]-\mathbb{E}[X(t<T)|A=0]\}
\end{aligned}
\end{equation}

A common solution to model the DiD causal effect is to specify a linear regression model for the observed outcome \cite{angrist2014mastering}. In case of one intervention and one treatment group we will have:
\begin{equation}
\begin{aligned}
            X = \alpha + \beta g + \gamma t + \delta (g\times t) + e,
\end{aligned}
\end{equation}
where $\alpha$, $\beta$, $\gamma$, and $\delta$ are the regression coefficients. $g$ is a dummy variable indicating the treatment group (1 for treatment and 0 for the control group), $t$ is a dummy variable defining the pre- or post-treatment periods (0 for before and 1 for after the treatment time), $g\times t$ is the interaction term to count the DiD causal effect with $\delta$ being its coefficient, and $e$ is the error term which is independent of other variables. The parameters of this linear regression can be estimated by the available methods solving regression models such as ordinary least squares (OLS) or gradient descent. Other variations of DiD regression consider settings with various treatments and multiple treatment groups \cite{angrist2014mastering, angrist2008mostly}, non-linearity assumption for DiD model \cite{athey2006identification}, and two control groups (known as triple differences) \cite{atanasov2016shock, wing2018designing}. For a comprehensive survey on this metric, we refer the readers to two studies provided by Lencher \cite{lechner2011estimation} and Bertrand et al. \cite{bertrand2004much}.  

Although in the design of the DiD, a temporal component is used, it has been pointed out that if it fits to highly auto-correlated data, the model underestimates the effect of the intervention \cite{bertrand2004much}. To overcome this problem, Brodersen et al. \cite{brodersen2015inferring} propose a method named \textit{Causal Impact} which is widely used for various applications such as the impact of vaccines, the environmental impact of aircraft emissions and aviation fuel tax, and the impact of mobile phone use on brain cancer \cite{bruhn2017estimating, gonzalez2016environmental, de2016inferring, de2017intervention}. This method generalizes the concept of the DiD and structural time-series model to infer the causal impact of a discrete intervention (e.g. the release of a new product). Causal impact learns the relationship between the treatment and control group before any intervention and predicts the counterfactual series after the treatment. This method relies on state-space models as follows \cite{samartsidis2019assessing}:
\begin{equation}
\begin{aligned}
            X(t) &= \mu(t)+\beta Z(t)+v(t)\\
            \mu(t) &= \mu(t-1)+\delta(t-1)+w(t)\\
            \delta(t) &= \delta(t-1)+u(t),
\end{aligned}
\end{equation}
where $Z(t)$ is the control time series and is related to the treatment time series (i.e. $X(t)$) through the $\beta$ components. $v(t)$, $w(t)$, and $u(t)$ are zero-mean noise variables and $\mu(t)$ models the temporal correlation in $X(t)$.  The $\delta(t)$ component can be thought of as the slope at time $t-1$ or the expected increase in $\mu$ between times $t-1$ and $t$.
The model is fitted to the observed data $t = 1, 2, ..., T$, treating the counterfactual $t= T+1, T+2, ..., n$ as unobserved random variables. With these, the model will compute the posterior distribution of the counterfactual time series. The causal effect is estimated by subtracting the predicted from the observed treated time series, which captures the semi-parametric Bayesian posterior distribution. Li and Bühlmann \cite{li2018estimating, li2018estimatingpaper} proposed a state-space-based model named \textit{Causal Transfer}, inspired by \textit{Causal Impact}, to learn the effect of the treatment in a time series data and capture how it evolves over time in order to transfer it to other time series. Specifically, after the treatment, we are only able to observe the outcomes under the treatment for one time series and under the control for another one, but not the potential outcome under control for the former and under treatment for the latter. The authors fill the missing outcomes by learning the intervention effect through a state-space model.

\textit{Synthetic Control Method} (SCM) introduced by Abadie and Gareazabal \cite{abadie2003economic} in 2003 overcomes the problem of ambiguities in the selection of control groups with the aim to estimate the effects of interventions that take place on an aggregate level (such as countries, regions, cities). In this method, we find the weight for each control unit such that the weighted average of all these potential control units (named as \textit{donor pool}) best resembles the characteristics of the treated unit before the treatment and use the learned weights to estimate the counterfactual after the intervention. Formally, the SCM finds the weights by minimizing:
\begin{equation}
\begin{aligned}
            W^*=\argmin_W||X_{a=1}(t<T)-X_{a=0}(t<T)W||\\
            \text{s.t. } \sum_{i=1}^J w_i=1 \text{ and } w_i\geq 1,
\end{aligned}
\end{equation}
 where $X_{a=1}(t<T)$ is a $T \times 1$ treated unit vector and $X_{a=0}(t<T)$ is a $T \times J$ matrix with $J$ being the number of control units. The predicted counterfactual of the treated unit is calculated by  
 \begin{equation}
\begin{aligned}
            X^c_{a=1}(t>T)=X_{a=0}(t>T)W^*
\end{aligned}
\end{equation}
 
 Other variations such as allowing for multiple treated units \cite{kreif2016examination} and applications of this model \cite{abadie2010synthetic, aytuug2017twenty, cavallo2013catastrophic, cole2020impact, mitze2020face, saunders2015synthetic} have also been conducted by researchers.

\begin{figure}[ht]
  \centering
  \includegraphics[width=0.90\linewidth]{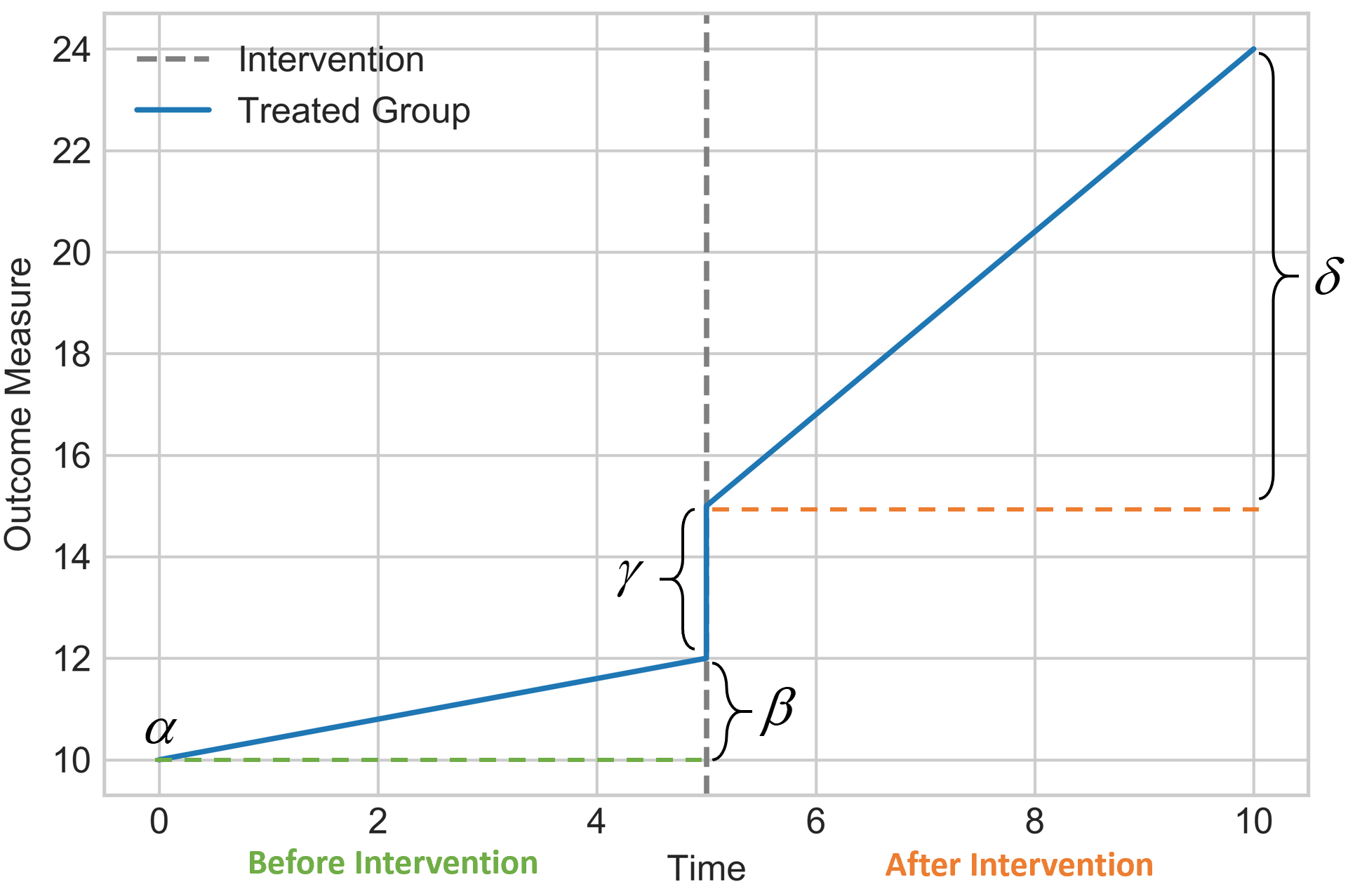}
  \caption{An illustration of the Standard ITS method.} 
  \label{fig::interrupted-TS} 
\end{figure}

In the case when (1) the intervention begins at a known point in time, (2) the outcome changes relatively faster after the intervention or a defined lag, and (3) the outcome lasts long enough to measure \cite{bernal2017interrupted}, a method name \textit{Interrupted Time Series} (ITS) analysis can be used. Although different variations of ITS exist, the standard ITS model is capable of finding the causal effect of an intervention for only one time series (i.e. when the control unit data is not available). This method is built upon a simple idea that the data generating process would have continued in a similar way in the absence of the intervention, which is a special case of \textit{Regression Discontinuity Design} (RDD) when the discontinuity happens in time \cite{hausman2018regression, kontopantelis2015regression}. Therefore, to find immediate changes in the outcome value and the change in the trend of the time series in the post-intervention period compared to the pre-intervention period, ``Segmented Regression'' is used \cite{penfold2013use}:
\begin{equation}
\begin{aligned}
            Y = \alpha + \beta T + \gamma t + \delta (T\times t) + e,
\end{aligned}
\end{equation}
where $\alpha$ indicates the baseline level at $T=0$, $\beta$ represents the baseline trend (i.e. pre-intervention trend), $\gamma$ is the immediate level change following the intervention, and $\delta$ indicates the post-intervention trend and $e$ is the error (Figure \ref{fig::interrupted-TS}).

Eichler and Didelez \cite{eichler2012causal} proposed a formal definition of causality along with an identifiability criteria for estimating the causal effects based on the intervention in dynamic Bayesian networks. The authors considered the effect of intervention in one component of the multivariate time series, at a specific point of the time, on another or the same component at later time points. They assumed that the causal effect excludes instantaneous causality and depends only on past variables. 
Using graphical models for the time series data, they proposed that if $(a,b)$ is not a directed edge in the graph, then the time series data of component $a$ at time $t$ has no causal effect on time series data of component $b$ at time $t+1$. Moreover, if $a$ is not the ancestor of $b$ in the graph, then the time series data of component $a$ at time $t$ has no causal effect of the component $b$ in the future. 


%% file: CTEEstimation/Time-varying.tex
\subsubsection{Time-varying Treatment Effect}
Most treatment effect estimation problems do not fit into the simple dichotomous treatment framework and require multiple sequential treatments which varies according to the time of the treatment~\cite{cooley2010identification}. For example, a drug dose when the dose is readjusted according to the patient's clinical response~\cite{time-var-expo}. In this case, $A$, the treatment variable, would be time-dependent and is recorded throughout all the time ($A(t)=\{A(1),A(2),...A(n)\}$). Below, we will briefly review the works in this area.

Works conducted by Robins et al.~\cite{robins1999estimation} as well as Hedeker and Gibbons \cite{hedeker2006longitudinal} expand the definitions of consistency, positivity, and conditional explainability assumptions along with the identifiability conditions of the time-invariant treatment effect estimation to estimate the causal effect of time-varying treatments. Moreover, in the presence of time-dependent counfounding, counterfactual inference of a time-dependent treatment has been extensively studied in the epidemiology literature specially in the chain of works conducted by Robin~\cite{hernan2000marginal, robins1992estimation, robins1997causal, robins2000marginal} on \textit{structural nested models} (SNM) and \textit{marginal structural modeling} (MSM). MSM, which is an alternative to SNM~\cite{robins2000marginal}, models the potential outcomes associated with each possible treatment trajectory with the \textit{inverse probability of treatment weighted} (IPTW) estimator~\cite{wodtke2020regression}. Specifically, a weight is assigned to each observation proportional to the inverse of the probability of treatment received given time-dependent confounders and previous treatments.
The prediction models of these methods are typically based on linear or logistic regression. Therefore, in case the outcome or the treatment policy exhibit complex dependencies on the covariate history, it would output an inaccurate result. To overcome this problem, Lim et al.~\cite{lim2018forecasting} proposed a deep learning method, \textit{Recurrent Marginal Structural Networks} (R-MSN), to learn time-dependent effects by using an RNN architecture for treatment response estimation based on the MSM framework. 
Wodtke~\cite{wodtke2020regression} suggests an alternative method for estimating marginal effects using \textit{regression-with-residuals} (RWR) estimation of a constrained structural nested mean models. Unlike IPTW, this method does not require a model for the conditional probability of treatment at each time point making it a good candidate for continuous-valued treatment problems.

Another category of approaches to capture the treatment effect of the longitudinal data with time-varying treatment is the \textit{Bayesian nonparametric} (BNP) methods. Xu et al.~\cite{xu2016bayesian} used a flexible Bayesian non-parametric approach to estimate the disease trajectories as well as the univariate treatment response curves from sparse observational time series. Inspired by this work, Soleimani et al.~\cite{soleimani2017treatment} used the flexible Bayesian semiparametric approach and linear time-invariant dynamical systems to model the treatment effects in multivariate longitudinal data by capturing the dynamic behavior (i.e. time-varying treatments). Their model composed of three components, one that captures the treatment response and the other two components models the natural evolution of the signal independent of the treatment.

Most of the aforementioned methods assume that all the confounders are observed. In other words, they consider that variables affecting the treatment assignments and the potential outcomes are all known, otherwise it will lead to a biased estimate of the outcome \cite{bica2019time,lok2008statistical}. A natural way to overcome this problem is through \textit{Factor Models} \cite{chan2016policy, gobillon2016regional, samartsidis2020bayesian, xu2017generalized}. A recent deconfounder method introduced by Bica et al. \cite{bica2019time} estimates the individual response to treatments in the presence of the multi-cause hidden confounders. To capture the distribution of the treatments over time a factor model was built to create the latent variables. Moreover, to make sure that this factor model is able to estimate the distribution of the assigned causes, a validation set of subjects were considered in order to compare the similarity of the two test statistics. These hidden confounders modeled by the latent variables were calculated by Recurrent Neural Network with multitask output and variational dropout. Similarly, in the presence of hidden confounders, Liu et al. \cite{liu2020estimating} proposed \textit{Deep Sequential Weighting} (DSW) for estimating the ITE with time-varying confounders. The authors used deep recurrent weighting neural network to combine the current treatment assignments with historical information and learn a new representation of hidden confonders to predict the potential outcome. 

%% file: CTEEstimation/DynamicRegimes.tex
\subsubsection{Dynamic Treatment Regimes}
\begin{figure}[ht!]
    \centering
    \includegraphics[width=0.7\linewidth]{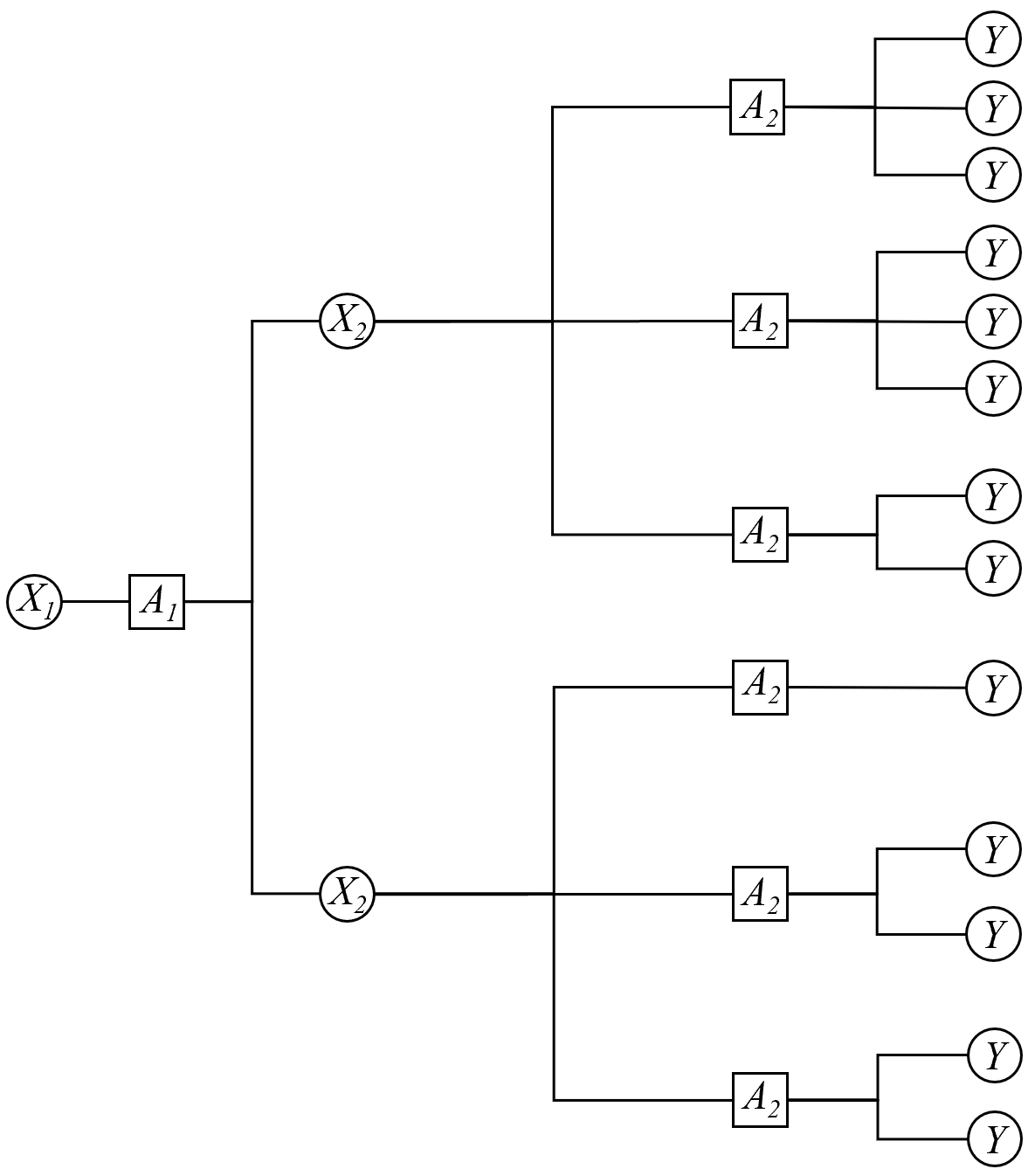}
    \caption{Dynamic treatment regimes for two intervals. $X$ and $A$ are categorical data, hence the tree-structured illustration~\cite{moodie2007demystifying}.}
    \label{fig:regime}
\end{figure}

With rapid-increasing interest in providing personalized treatment suggestions, dynamic regimes treatments are designed to provide treatment to individuals only when they need the treatment. A dynamic treatment regime is a function which takes in treatment and covariate history as arguments and outputs an action to be taken, providing a list of decision rules for how treatment should be allocated over time. 
Figure~\ref{fig:regime} shows a two-stage dynamic treatment regimes, where $X$ and $A$ denote the categorical covariates and the treatment, respectively. The observable data trajectory for a participant in a two-stage treatments is denoted by ($X_1,A_1,X_2,A_2$), where $X_1$ is the pre-treatment covariates and $X_2$ is the time-varying covariates which may depend on treatment received in the ﬁrst interval. The randomized treatment actions are $A_1$ and $A_2$ and the primary outcome is $Y=f(X_1,A_1,X_2,A_2)$.
For example, $X_2(A_1)$ denote a person's potential covariate status at the beginning of the second interval if treatment $A_1$ is received by that person and $Y$ denote the potential outcome if follows regime $(A_1,A_2)$.

\textit{Regret function} is widely used in dynamic treatment regimes to estimate the effect by the large scale treatments. To learn the parameters $\psi$  in regret function, Moodie et al.~\cite{moodie2007demystifying} use g-estimation, which is proposed by Robins~\cite{robins2004optimal}. For the purpose of estimation, $S_j(a_j)=s_j(a_j,h_j)$ depends on variables which are considered as interaction with treatment to influence outcome, where $h_j$ is unmeasured confounders. 
For example,
if the function at the second interval is linear,  $\gamma_2(h_2,a_2)=a_2(\psi_0+\psi_1x_1+\psi_2a_2+\psi_3x_2a_1)$, then the analyst may choose $S_2(a_2)=\frac{\partial}{\partial \psi}\gamma_2(h_2,a_2)=a_2(1,x_2,a_1,x_2a_1)^T$. Let 
\begin{equation}
    U(\psi,s) = \sum_{j_1}^{2} h_j(\psi){S_j(a_j)-E[S_j(a_j)|H_j]}
\end{equation}
with the probability of being the treated model. $E[U(\psi,s)]=0$ is an unbiased estimating equation from which consistent estimates $\hat{\psi}$ of $\psi$ may be found.

To  avoid estimating the full multivariate distribution of the longitudinal data, Murphy et al.~\cite{murphy2003optimal} model the multivariate distribution with two groups of parameters to estimate optimal rules. The first group of parameters (parameters in the regret functions) will be estimated and used to estimate the optimal rules. The the second group of parameters (most of which are infinite dimensional) are nuisance parameters.
Zhu et al.~\cite{zhu2020causal} extend  Boruvka et al.'s~\cite{boruvka2018assessing} definition of lagged treatment effect to continuous treatments to define a weighted advantage function and proposed a novel SNM.







%% file: Evaluation/evaluation.tex
\section{Performance Evaluation}
\label{eval}


In this section, we present an overview of the benchmark datasets and metrics used in time series and causal time series literature.

The datasets and metrics that are used for evaluation are based on the type of models that we are evaluating . The fact that causal time series analysis targets a different problem than traditional time series modelling, means that there are a lot more datasets available for time series which do not have causal metadata.
\input{Evaluation/datasets1}
\input{Evaluation/metrics1}

%% file: Evaluation/datasets1.tex
\subsection{Datasets}

In this section, we briefly introduce some of the datasets used in causal inference literature on time series. 
First, we begin by introducing datasets for the traditional time series forecasting. The functionalities of these datasets cannot be exploited for learning causality in time series as some datasets may lack ground truth information or may lack other features necessary for causal inference. To deal with these issues we introduce datasets specifically catered for the causal discovery problem as well as the treatment effect estimation problems.

\subsubsection{Time Series Datasets}
Traditionally, time series analysis uses a variety of data. In this section, we discuss some of the most commonly used datasets for traditional time series problem.
\begin{itemize}
    \item \textbf{UCR Time Series Classification Archive}: The UCR Time Series Classification archive consists of more than 120 datasets. These datasets, represent a classification problem. Thus, there is a class label for each item in every dataset. The UCR time series dataset is used in various publications \cite{fawaz2019deep,bagnall2015,xing2010brief}.
    \item \textbf{Baydogan's Archive}: Baydogan's Archive consists of over a dozen diverse multivariate time series datasets from diﬀerent applications such as speech recognition, activity recognition, medicine and etc. This dataset is used and introduced in \cite{baydogan2015learning}.
    \item \textbf{NYC Taxi Dataset}: This is a univariate time-series dataset containing the New York City (NYC) taxi demand from 2014-–07–01 to 2015–01–31 with an observation of the number of passengers recorded every half hour containing 10320 timestamps. This dataset is recommended by Braei et al. \cite{braei2020}.
    \item \textbf{Real Yahoo Services Network Traffic Dataset}: This is a univariate time-series dataset containing the traffic to Yahoo services. The anomalies are labeled by humans. This dataset consists of 67 different time-series each containing about 1400 timestamps. This dataset is recommended by Braei et al. \cite{braei2020}.
    \item \textbf{Synthetic Yahoo Services Network Traffic Dataset}: This dataset consists of 100 synthetic univariate time-series data containing anomalies. Each time-series contains about 1421 timestamps. The anomalies have been inserted randomly therefore representing point anomalies. This dataset is recommended by Braei et al. \cite{braei2020}.
\end{itemize}

As mentioned earlier, these datasets are fruitful when it comes to traditional time series problems, like classification and forecasting. For inferring causality, one may need access to treatment and control groups for treatment effect estimation and may need access to variables such as confounders to interpret correct causal relations between different variables. Thus, the traditional datasets cannot be used for inferring causality as they lack these features. We now introduce datasets relevant for treatment effect estimation and causal discovery problems. 

\begin{table*}[ht]
	\centering
	\begin{tabular}{|p{0.2\textwidth} | p{0.3\textwidth} |p{0.3\textwidth} |}
		\hline
		 \textbf{Model}       & \textbf{Description}                                        & \textbf{Datasets}                                                                                                         \\ \hline
		 Confounding             & Multiple effects from the same cause                        & Bica et al. \cite{bica2019time}                                                                                          \\ \hline
		 Non-linear           & Absence of a linear relationship between cause and effect   & Financial Time-Series \cite{kleinberg2013causal}, Particle Simulation \cite{lowe2020amortized}                            \\ \hline
		Dynamic Time dependence & Dependence on the time-lagged component(s) varies over time & Li et al. \cite{li2018estimating}                                                                                        \\ \hline
		Chaotic              & Small changes in parameters create large changes over time  & Papana et al. \cite{papana2013simulation}, Khanna et al. \cite{khanna2019economy}                                       \\ \hline
	\end{tabular}
    
\caption{Properties of synthetic datasets used in literature}
\label{tbl:synthetic}
\end{table*}

\subsubsection{Treatment Effect Estimation Datasets}
In this section, we discuss some of the commonly used real world datasets for the treatment effects estimation problem.
\begin{itemize}
    \item \textbf{MIMIC II/III Data} \cite{johnson2016mimic}: This dataset consists of data about patients in ICU. Various attributes about the patients are stored in this database. Examples include blood pressure, oxygen saturation, given medicine, as well as temporal attributes. Thus, it is a gold-mine for causal research.This dataset has been used by Bica et al. \cite{bica2019time} to estimate treatment effects. The dialysis subset of this data has been used by Soleimani et al. for Treatment-Response modeling using counterfactual reasoning \cite{soleimani2017treatment}.
    \item \textbf{Advertisement Data}: Brodersen et al. \cite{brodersen2015inferring} use advertisement data by Google to determine the causal impact. An advertisement campaign was analyzed by the authors where the product ads were placed alongside search results for a period time of 6 weeks. The ad data was used as an intervention to measure the impact on sale volume as the effect.
    \item \textbf{Geo experiment data} \cite{kerman2017estimating}: This data consists of an ad campaign as a treatment for half of the non-overlapping geo data in the sample. It was used by \cite{li2018estimating} for experiments in their approach.
    \item \textbf{Economic data for Spanish regions}: In a case study about the economic cost of conflict \cite{abadie2003economic} use economic data for Spanish regions to analyze the effects of terrorism. The authors use the per capita GDP of Basque over time to do their causal analysis. The data does not have a ground truth value.
    \item \textbf{California's Tobacco Control Program}: Another work looks at the effects of California’s tobacco control program in their work using Synthetic Control Methods \cite{abadie2010synthetic}.  Here they use annual state-level panel data. This data contains the per-capita cigarette sales from 1970-2000 for multiple US states. This is the time between which Proposition 99 was passed.
    \item \textbf{Air Quality Data} \cite{auffhammer2011clearing}: This dataset is used to study the effects of gasoline content on air quality. This dataset includes Ozone levels, minimum/maximum/mean temperatures, precipitation, and snow information. The data goes over the years 1989 till 2006.
    \item \textbf{Monetary Policy Data}: This dataset comes from three different sources. Quarterly GDP for Switzerland and Euro are taken from Eurostat. The monthly business confidence index and monthly consumer price index for Switzerland are taken from OECD. Monthly balance sheet data, monthly call money rate, and monthly average exchange rate are taken from the Swiss National Bank. Pfister et al.~\cite{pfister2019invariant} use this dataset to satisfy their goal of finding monthly causal predictors for log-returns of the Euro-Swiss franc exchange rate. The authors use data from the year 1999 to 2017.
\end{itemize}

\subsubsection{Causal Discovery Datasets}
Datasets for causal discovery range from economical data \cite{gong2017causal} to health data \cite{schaechtle2013multi}. We discuss some of the commonly used real world datasets.
\begin{itemize}
    \item \textbf{US Manufacturing Growth Data}: This dataset consists of microeconomic data of growth rates of US manufacturing firms in terms of employment, sales, research \& development (R\&D) expenditure, and operating income, for the years 1973–2004. It can be used to identify the causal variables that affect the growth rate of a firm. It has been used in \cite{entner2010causal}.
    \item \textbf{Diabetes Dataset}: This dataset consists of Diabetes patient records that were obtained from two sources: an automatic electronic recording device and paper records. \cite{schaechtle2013multi} use it to deduce the ground truth causal graph.
    \item \textbf{Temperature Ozone Data}: This dataset consists of two variables, 72 points in time, 16 different places. The two variables are ozone and radiation with the assumed ground truth that radiation has an causal effect on ozone. This dataset was used by \cite{schaechtle2013multi,gong2017causal,mooij2016distinguishing}.
    \item \textbf{OHDNOAA Dataset}: This is a dataset by the Office of Hydrologic Development at the National Oceanic and Atmospheric Administration which consists of 32 hydrology related variables over several square areas for the US. The data is collected at constant intervals of 6 hours and ranged from the years 1979 to 2008. It is used by \cite{jangyodsuk2014causal}.
    \item \textbf{Neural activity Dataset}: This dataset consists of real-time whole-brain imaging to record the neural activity of Caenorhabditis elegans. The dataset consists of 302 neurons and is generally used to identify which neurons are responsible for movement.
    \item \textbf{Human Motion Capture}: This dataset is from CMU MoCap database contains data about joint angles, body position from two subjects. The dataset contains 54 joint angles over 2024 time points. Tank et al. \cite{tank2018neural} use this dataset in their Causal Discovery work. 
    \item \textbf{Traffic Prediction Dataset}: This dataset contains four months’ worth of sensor data from Los Angeles, California. 207 sensors are placed for collecting this data. The location of each sensor in the form of GPS coordinates are also included in the dataset. It is used by \cite{pan2018hyperst}.
    \item \textbf{Stock Indices Data}: Stock Indices are a source of data used in Causal work. Rothenhausler  et  al. use NASDAQ, S\&P 500, and DAX indices for a period between 2000-2012 for their Causal Discovery work \cite{rothenhausler2015backshift}. They create 74 blocks of data. Each block represents 61 consecutive days.
\end{itemize}

\subsubsection{Synthetic Datasets}
Even though the real world datasets exist to solve the causal discovery and the treatment effect estimation problems, many researchers use synthetic datasets for the purpose of illustrating particular technical difficulties inherent to some causal models e.g. Markov equivalence (several causal graphs are consistent with the same data). The truth values of causal relationships are known for such data generating models. In the following section we discuss some synthetic dataset generation methods and some works that use them. A summary of the synthetic models is presented in Table \ref{tbl:synthetic}.

\begin{itemize}
     \item \textbf{Confounding/ Common-cause Models}: One of the concerns in causal literature is the effect or discovery of confounders. Thus, in order to model them, there are several approaches in the literature. Huang et al. use simulated datasets with a common cause and common effect \cite{huang2015fast}. The datasets contain noise variables and causal variables over time. Let $e(t)$ be the value of a variable $e$ at time $t$. Let $c$ represent the variable which causes $e$. Then:
     $$e(t)=\sum_{c\in X}\sum_{i=1}^{n}I(c, e)$$ where $n=|T(c)\cap[t-s, t-r]|$ and $I(c, e)$ is the impact of $c$ on $e$.
    \item \textbf{Non-Linear Models}: Since many real-world systems are non-linear, there are works which try to simulate non-linear systems to evaluate their approach. An example of non-linear model is the simulation used by Papana et al.~\cite{papana2013simulation}. In their work, they simulate a tri-variate system with linear($X_2 \rightarrow X_3$) and non-linear($X_1 \rightarrow X_2$, $X_1 \rightarrow X_3$) relationships:
    \begin{align*}
   	x_{1,t} &= 3.4 x_{1, t-1} (1-x_{1, t-1})^2 \exp(-x_{1,t-1}^2)+0.4\epsilon_{1,t} \\
   	x_{2,t} &= 3.4 x_{2, t-1} (1-x_{2, t-1})^2 \exp(-x_{2,t-1}^2)+0.5x_{1,t-1}x_{2,t-1} \\ &{\ \ \ }+ 0.4\epsilon_{2,t}\\
   x_{3,t} &= 3.4x_{3,t-1}(1-x_{3,t-1})^2exp(-x_{3,t-1}^2)+0.3x_{2,t-1}\\&{\ \ \ }+0.5x_{1,t-1}^2+0.4\epsilon_{3,t}
    \end{align*}
    where $\epsilon_{i,t}$, $i=1,...,3$ are Gaussian white noise processes.
    There are various other works which use non-linear models of data~\cite{lowe2020amortized}.

     \item \textbf{Dynamic Models}: Some models try to simulate a situation where the dependence of the variables vary over time in a non-linear and non-exponential manner. An example is the model used by Li et al.~\cite{li2018estimating} where there is a sinusoidal dependence:
     \begin{multline*}
     X_t = \cos(X_{t-1} + X_{t-4}) + \log (|X_{t-6} - X_{t-10}| + 1) + \epsilon_t
     \end{multline*}
     For treatment effects, one of the examples of a model used in literature is defined as follows~\cite{kerman2017estimating}:
     \begin{multline*}
	X_{i,t} = \beta_{0,t} + \beta_{1,t}X_{i,pre} +\beta_{2,t} Z_{i,t} + T_i\mu_{i,t}\cos(X_{i,pre}) +v_{i,t}
	\end{multline*}
	where $Z$ is the time-varying covariate, $v_{i,t}$ is the noise, $X_{i,pre}$ is the pre-treatment covariate, $T_i$ is the treatment indicator and $\mu_{i,t}$ represents the state.
	
    \item \textbf{Chaotic Models}: Choaticity is the ability of a model to deviate over different values of its hyperparameters. Chaoticity is usually expressed in terms of a Lorenz model. One of the examples of such a model which is also non-linear with $X_1 \rightarrow X_2$ and $X_2 \rightarrow X_3$ can be expressed as follows~\cite{papana2013simulation}:
    \begin{align*}
    \dot{x}_1 &= 10(y_1 - x_1) \\
    \dot{x}_2 &= 10(y_2-x_2) + c(x_1 - x_2) \\
    \dot{x}_3 &= 10(y_3 - x_3) + c(x_2 - x_3) \\
    {\ }\\
    \dot{y}_1 &= 28x_1-y_1-x_1z_1 \\
    \dot{y}_2 &= 28x_2-y_2-x_2z_2 \\
    \dot{y}_3 &= 28x_3-y_3-x_3z_3 \\
    {\ }\\
    \dot{z}_1 &= x_1 y_1 - 8/3 z_1 \\
    \dot{z}_2 &= x_2 y_2 - 8/3 z_2 \\
    \dot{z}_3 &= x_3 y_3 - 8/3 z_3
    \end{align*}
    where $c$ is the coupling strength. Variations of the Lorenz model is used in other places in causal inference for time series. The Lorenze-96 model is used in the work by Khanna et al.~\cite{khanna2019economy} where the authors mention it as a popular model for climate science.
\end{itemize}

%% file: Evaluation/metrics1.tex
\subsection{Evaluation Metrics}
In  this  section, we go over different metrics used in the literature for measuring how well a model performs. We begin by covering the metrics used for traditional time series problems like forecasting and classification. We then move on to metrics designed for causal discovery followed by the metrics for treatment effect estimation problem. 

\begin{table*}[!ht]
\begin{center}

\begin{tabular}{| >{\centering\arraybackslash}m{0.25cm}|m{2cm}|m{4.8cm}| >{\centering\arraybackslash}m{5.2cm}| >{\centering\arraybackslash}m{0.25cm}| >{\centering\arraybackslash}m{0.25cm}| >{\centering\arraybackslash}m{0.25cm}| >{\centering\arraybackslash}m{1.7cm}|} 
\hline
   & \textbf{Metric Name}   & \textbf{Notions} & \textbf{Definition} & \rotatebox[origin=c]{90}{\textbf{Correctness}} & \rotatebox[origin=c]{90}{\textbf{Comparative}} & \rotatebox[origin=c]{90}{\textbf{Significance}} & \textbf{Commonly Used in} \\ 
\hline\hline

1  &  TPR & \multirow{2}{=}{{Let $G = (V, E_M)$ be the causal graph inferred by the model, where $V_M = \{v_1, v_2, ..., v_n\}$ represent the covariates (including time-lagged) and $E_M$ represent the edges between the covariates if there is a direct causal dependence. Similarly, let $G' = (V, E_{GT})$ represent the covariates and the edges in the ground truth. Let $E = \{e_1, e_2, ..., e_m\}$ represent all possible edges between $V$.}} &    $TPR = \mathlarger{\sum}_i \frac{e_i}{|E_{GT}|},\ e_i \in E_M \cap E_{GT}$     &     \cmark         &            &             &  Causal Discovery             \\[5.5em]
\cline{1-2}\cline{4-8}
2  & FPR        &         &  $FPR = \mathlarger{\sum}_i \frac{e_i}{|E - E_{GT}|},\ e_i \in E_M \backslash E_{GT}$      &   \cmark          &            &             & Causal Discovery              \\ [5.5em]
\hline
3  & MSE           &   Let $T$ be the inferred transition matrix and $A$ be the adjacency matrix for the ground truth.        &   $MSE = \frac{1}{|T|} \sum (T-A)^{\circ 2} $      &   \cmark          &            &             & Both              \\ 
\hline
6  & Precision     &   Let $TP$ and $FP$ denote the True Positive and False Positive, respectively.         &   $	Precision = \frac{TP}{TP + FP} $      &     \cmark        &            &             & Causal Discovery\\ 
\hline
7  & Recall        &     Let $TP$ and $FN$ denote the True Positive and False Negative, respectively.       &     $	Recall = \frac{TP}{TP + FN}	$    &     \cmark        &            &             & Causal Discovery\\
\hline
8  & F1-Score      &     Let $P$ and $R$ be the precision and recall, respectively       &    $	F_1 = \frac{2PR}{P + R}	$    &   \cmark          &            &             & Causal Discovery               \\ 
\hline
9  & F-Test        &    Let $RSS_0 = \sum_{t=1}^T \hat{e}_t$ and $RSS_1 = \sum_{t=1}^T \hat{u}_t$, where T is the length of the time series, $\hat{e}_t$ and $\hat{u}_t$ is the time dependent error for the null hypothesis and the alternative hypothesis, respectively.     &  $F = \frac{(RSS_0 - RSS_1)/p}{RSS_1/(T-2p-1)}$       &             &   \cmark         &   \cmark          & Both              \\ 
\hline
10 & Unpaired T-Test        &    Let $\bar{x}_1$ and $\bar{x}_2$ be the means of two sequences. Let $s_1$ and $s_2$ represent the standard deviation of the sequences. Let $n_1$ and $n_2$ represent the cardinalities of the sequences.        &     $utt = \frac{\bar{x}_1 - \bar{x}_2}{\sqrt{(\frac{1}{n_1}+\frac{1}{n_2})(\frac{(n_1-1)s^2_1 + (n_2-1)s^2_2}{(n_1-1)+(n_2-1)})}} $    &             &            &   \cmark          &  Treatment Effect Estimation             \\
\hline
\end{tabular}
\caption{Causal time series metrics and their definition.}
\label{tab:causal}
\end{center}
\end{table*}

\subsubsection{Time Series Metrics}
There are multiple time series metrics used in the literature \cite{bagnall1602great,fawaz2019deep, lines2016hive}. In the following we discuss some of the most common ones.
\begin{itemize}
    \item \textbf{Accuracy}: One of the common metrics for validating time series classification models is accuracy. The accuracy is simply the percentage of samples that are correctly classified.\cite{fawaz2019deep, lines2016hive}. 
    \item \textbf{Mean/Median Error}: There are several variations of errors used in time series literature, particularly for time series forecasting~\cite{hyndman2006another} such as (Root) Mean Squared Error, Mean/Median Absolute Error, Mean/ Median Absolute Percentage Error, Symmetric Mean/ Median Absolute Percentage Error, Median Relative Absolute Error, Geometric Mean Relative Absolute Error, and Mean Absolute Scaled Error. All of these are intended to measure the differences between the forecasted time series and the ground truth.
    \item \textbf{Longest Common SubSequence}: The LCS \cite{bagnall1602great} measures the similarity of two time series with a matching threshold $\theta$. A threshold $\theta$ is used to determine the state of match or mismatch between two time series. If the Euclidean distance of two data points is not greater than $\theta$, then the two data points are considered to be a match, else they are said to mismatch.
    \item \textbf{Edit Distance with Real Penalty}: In this metric, the distance between the two series is measured by looking at the number of operations required to change one series into the other series. A distance matrix is created which stores the distances between the two time series across multiple time points inside the time window. Gaps between points are penalized according to a user-provided reference value~\cite{chen2004marriage}.
    \item \textbf{Euclidean Distance}: One of the simplest evaluation metrics for time series is the Euclidean Distance (ED) \cite{iglesias2013analysis}. Given two series, the square root of the sum of squared distances between each time step is the euclidean distance. It helps define the similarity between two independent time series and is suitable for applications that do not present direct or necessary correlation among distinct features.
    \item \textbf{Dynamic Time Warping}: The dynamic time warping (DTW) \cite{berndtDTW1994} is a metric for calculating the distance between two time series sequences. Suppose two sequences of time series $A = a_1,a_2 \dots ,a_m$ and $B = b_1,b_2, \dots b_n$ are given then, let DTW(i, j) denote the DTW distance of $A_{1..i}$ and $B_{1..j}$ . The distance function dis($a_{i}, b_{j}$ ) represents the distance of $a_i$ and $b_j$. The Euclidean distance is usually used to measure the distance between $a_i$ and $b_j$.
\end{itemize}

\subsubsection{Causal Time Series Evaluation Metrics}
In this section we discuss different metrics for the evaluation of causality in time series. We first, introduce metrics for the causal discovery problem followed by the metrics for the causal treatment effect estimation. A summary of the metrics can be found in Table \ref{tab:causal}.


\paragraph{Treatment Effect Estimation Metrics}
When we talk about metrics for treatment effect estimation, we refer to those metrics that give feedback on how successful a model was in estimating a specific value. For causal effect estimation, the metrics which are used, give us information about how well the estimation compares to the ground truth.
\begin{itemize}
    \item \textbf{MSE}: MSE and its variations (Root Mean Squared Error~\cite{pan2018hyperst}, Normalized Root Mean Squared Error~\cite{soleimani2017treatment}) are used commonly in causal effect estimation literature~\cite{bica2019time,kerman2017estimating}. Unlike MSE for causal discovery, here MSE is used to compare the inferred series with the ground truth series by taking the average of the squared differences at each time step.
    \item \textbf{F-Test}: F-test are used in treatment effect estimation to asses treatment effect heterogeneity by examining the marginal variances of the treatment and control outcomes. It is defined as the ratio of variance for the treated group over the variance for the control group.
    \item \textbf{T-Test}: The t-test is a metric used to compare two sequences. In their work on causal inference with rare time series data, Kleinberg et al.~\cite{kleinberg2013causal} use an unpaired t-test to determine the significance of a cause within a certain time-range. 
\end{itemize}

\paragraph{Causal Discovery Metrics}
Here we explain different metrics used in the literature for causal discovery. 
causal discovery mainly focuses on finding the causal relationships~\cite{entner2010causal, runge1702detecting}, so the metrics involved in this type of approach usually provide measures of correctly identified relationships. Below, we list some of the commonly used metrics for this task:
\begin{itemize}
    \item \textbf{Structural Hamming Distance}: Structural Hamming Distance (SHD) is a metric used to compare a discovered causal graph with the ground truth. More specifically, SHD describes the number of changes that need to be made to a graph to turn it into the graph its being compared with. This is calculated by counting the missing edges, extra edges and incorrect edge direction between two graphs. To asses the performance of causal discovery methods, SHD takes as input two partially directed acyclic graphs (PDAGs) and outputs the count of edges that do not coincide using the aforementioned process.
    \item \textbf{True/False Positive Rate}: The True Positive Rate for both adjacencies (discovered neighbors) and arrowheads (direction of discovered causal relations) are defined as the ratio of common edges found in the predicted and the ground truth adjacencies over the number of edges in the ground truth graph. It take as input the predicted adjacency matrix and the ground truth adjacency matrix to calculate the ratio. Similarly, the False Positive Rate refers to the ratio of common edges found in the predicted and ground truth adjacencies over the absolute difference of number of edges between ground truth and predicted adjacencies. These metrics have been used in  \cite{hyttinen2016causal,runge1702detecting,runge2018causal,schaechtle2013multi}.
    \item \textbf{Area Under the Receiver Operator Curve}: This measure is one of the most popular metrics for Causal Discovery. The Receiver Operator Characteristic (ROC) is defined as a ratio of True Positive Rate (TPR) and False Positive Rate (FPR). The ROC curve is created by iterating over the cut off for classification and recording the TPR against the FPR. The area under the ROC curve (AUROC) is then used to asses the performance of the model. The higher
    the value of this metric, the better the model is. This metric has been widely used in different papers such as \cite{haufe2010sparse,khanna2019economy, lowe2020amortized,wunonlinear,xu2019scalable}. 
    For Instance, in \cite{khanna2019economy}, the ROC curve illustrates the trade off between the TPR and the FPR achieved by the different methods towards the detection of pairwise Granger causal relationships between the $n$ measured processes in their experiment. 
    \item \textbf{Mean Squared Error}: This metric is used to evaluate works in causal discovery and for example, in Temporally Aggregated Time Series~\cite{gong2017causal}, the authors construct a transition matrix to represent the causal graph. Here, the MSE between the distance from the transition matrix to the ground truth is used to evaluate the causal relationships inferred by the proposed model.
    \item \textbf{F-Score, Precision and Recall}: Precision is defined as the ratio of correctly predicted positive observations to the total predicted positive observations. Recall is defined as the ratio of correctly predicted positive observations to the all observations in the actual class. F1 Score is defined as the weighted average of Precision and Recall. They are commonly used to evaluate the performance of a model. It take as input the predicted adjacency matrix and the ground truth adjacency matrix to calculate the true positives, false negatives, true negatives and false negatives.
    \item \textbf{Area Under the Precision Recall Curve}: Area Under the Precision Recall Curve (AUPR) is another metric used in literature~\cite{tank2018neural,xu2019scalable}. This metric relies on precision and recall. Similar to the AUROC, this metric measures the area under the Precision-Recall curve.
    \item \textbf{F-Test}: One approach to finding whether causality exists is the F-test. To use this test the user first defines a null hypothesis, for example in \cite{papana2013simulation} the null hypothesis is, the coefficients of the lagged driving variables in the unrestricted VAR model are all zero . Then, we can construct a restricted and unrestricted equation and estimate the hyper-parameters using Ordinary Least Squares. \begin{definition}[F-Test]
    Let $X = \{x_t, x_{t-1}, x_{t-2}, ... \}$ be one series. Let $Y = \{y_t, y_{t-1}, y_{t-2}, ... \}$ be another series. The null hypothesis is given by the restricted equation:
    \begin{equation}
    x_t = c_1 + \sum_{i=1}^p \gamma_ix_{t-i} + e_t
    \end{equation}
    Where $p$ is the lag, $c_1$ is the intercept and $e_t$ is the time-dependent error.
    The unrestricted equation represents the alternative hypothesis:
    \begin{equation}
     x_t = c_2 + \sum_{i=1}^p \alpha_ix_{t-i} + \sum_{i=1}^p \beta_i y_{t-i} + u_t
    \end{equation}
    Where $p$ is the lag, $c_2$ is the intercept and $u_t$ is the time-dependent error. Let $RSS_0 = \sum_{t=1}^T \hat{e}_t$ and $RSS_1 = \sum_{t=1}^T \hat{u}_t$, where T is the length of the time series. The F-test statistic is defined as:
    \begin{equation}
    F = \frac{(RSS_0 - RSS_1)/p}{RSS_1/(T-2p-1)}
    \end{equation}
    \end{definition}
\end{itemize}